\documentclass[journal]{IEEEtran}

\usepackage{cite}

\usepackage{amsmath,amssymb,amsfonts}
\usepackage{algorithmic}
\usepackage{graphicx}
\usepackage{subcaption}
\usepackage{booktabs}   
\usepackage{multirow}

\usepackage[linesnumbered,ruled,vlined]{algorithm2e}
\def\*#1{\mathbf{#1}}

\usepackage{textcomp}
\usepackage{soul}
\usepackage{bm}
\usepackage{makecell}
\usepackage{url}
\usepackage{xcolor}
\def\etal{\emph{et al.}}
\begin{document}
\title{Investigating and Improving Latent Density Segmentation Models for Aleatoric Uncertainty Quantification in Medical Imaging}
\author{M. M. Amaan Valiuddin, Christiaan G. A. Viviers, Ruud J. G. van Sloun, \textit{Member, IEEE}, Peter H. N. de With, \textit{Fellow, IEEE}, and Fons van der Sommen, \textit{Member, IEEE}

% \thanks{This paragraph of the first footnote will contain the date on which
% you submitted your paper for review. It will also contain support information,
% including sponsor and financial support acknowledgment. For example, 
% ``This work was supported in part by the U.S. Department of Commerce under Grant BS123456.'' }
% \thanks{Submitted for review on the 31st of July.}
% \thanks{The next few paragraphs should contain the authors' current affiliations,
% including current address and e-mail. For example, Authors are with the Department of Electrical Engineering
% Eindhoven University of Technology, Eindhoven, 5612 AZ, The Netherlands (e-mail:m.m.a.valiuddin@tue.nl). }
\thanks{M.M.A. Valiuddin, C.G.A. Viviers, R.J.G. van Sloun, P.H.N. de With and F. van der Sommen are with the Department of Electrical Engineering at the Eindhoven University of Technology, Eindhoven, 5612 AZ, The Netherlands (corresponding author e-mail: m.m.a.valiuddin@tue.nl). \\ \indent R.J.G. van Sloun is also with Philips Research, Eindhoven, The Netherlands.}}% \thanks{}
% \thanks{T. C. Author is with the Electrical Engineering Department,
% University of Colorado, Boulder, CO 80309 USA, on leave from the National
% Research Institute for Metals, Tsukuba, Japan (e-mail: author@nrim.go.jp).}}

%TODO: add lipschitz weight clipping thingy

\maketitle

\begin{abstract}
Data uncertainties, such as sensor noise, occlusions or limitations in the acquisition method can introduce irreducible ambiguities in images, which result in varying, yet plausible, semantic hypotheses. In Machine Learning, this ambiguity is commonly referred to as aleatoric uncertainty. In image segmentation, latent density models can be utilized to address this problem. The most popular approach is the Probabilistic U-Net (PU-Net), which uses latent Normal densities to optimize the conditional data log-likelihood Evidence Lower Bound. In this work, we demonstrate that the PU-Net latent space is severely sparse and heavily under-utilized. To address this, we modify the training objective to introduce the entropy-regularized Sinkhorn Divergence in the latent space, which promotes homogeneity across all latent dimensions, effectively improving gradient-descent updates and latent space informativeness. Applied to public datasets of various clinical segmentation problems, our proposed methodology achieves the best Empirical Wasserstein Distance on every dataset compared against preceding latent variable models for probabilistic segmentation. The results indicate that encouraging a homogeneous latent space significantly improves latent density modeling for medical image segmentation.
\end{abstract}

\begin{IEEEkeywords}
Probabilistic Segmentation, Aleatoric Uncertainty, Latent Density Modeling
\end{IEEEkeywords}

\section{Introduction}
\label{sec:introduction}
\begin{figure}[b]
    \centering
    \includegraphics[width=0.45\textwidth]{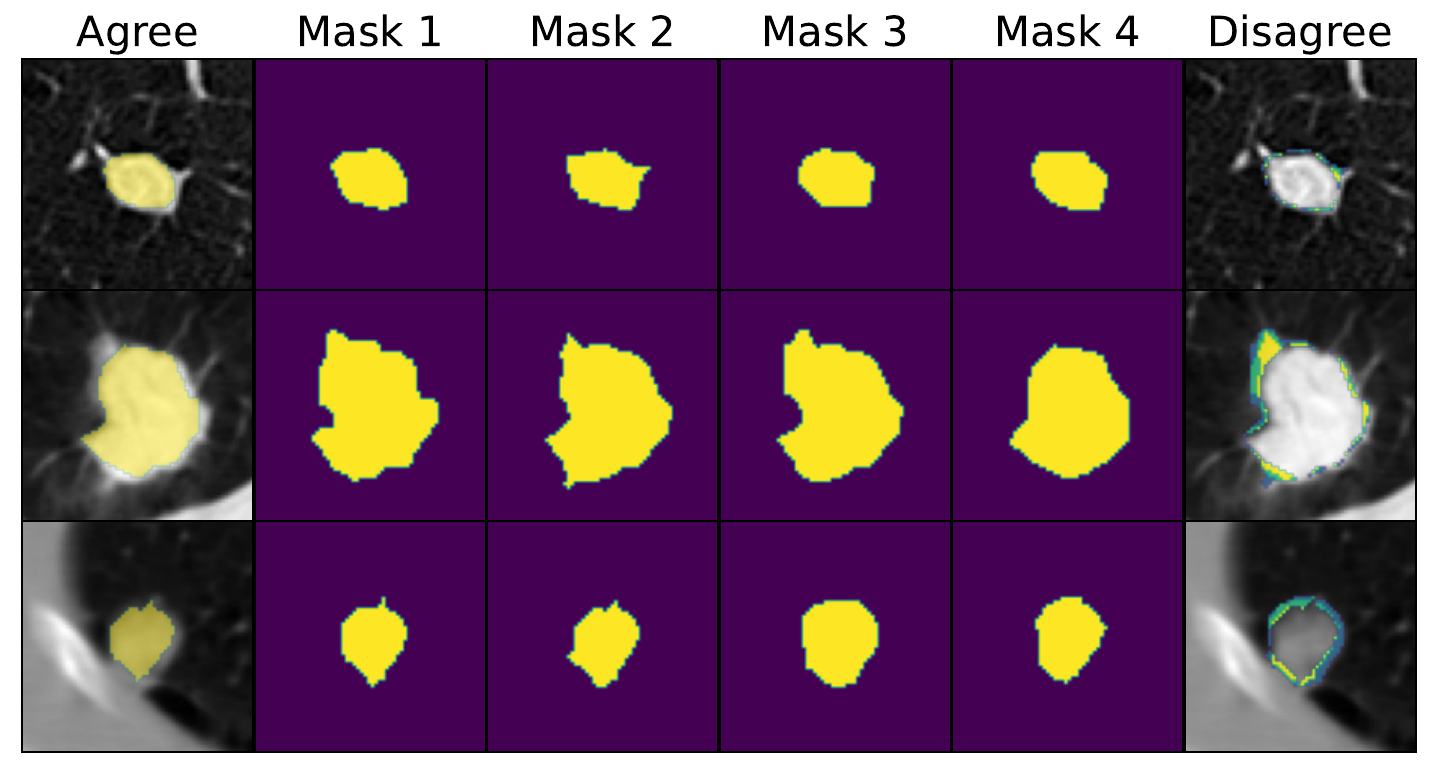}
    \caption{Samples of the LIDC-IDRI dataset with significant inter-observer variability where (dis-)agreement in the ground-truth masks is clearly visible.}\label{fig:lidcex}
\end{figure}

Supervised deep learning segmentation algorithms rely on the assumption that the provided reference annotations for training reflect the unequivocal ground truth. Yet, in many cases, the labeling process contains substantial inconsistencies in annotations by domain-level experts. These variations manifest from inherent ambiguity in the data (e.g., due to occlusions, sensor noise, etc.), also referred to as the \textit{aleatoric uncertainty}~\cite{kendall2017uncertainties}. In turn, subjective interpretation of readers leads to multiple plausible annotations. This phenomenon is generally expressed with multi-annotated data, revealing that labels may vary within and/or across annotators (also known as the \textit{intra-/inter- observer variability}). While this concept is generic, the most significant impact of this phenomenon can be encountered in medical image segmentation (see Figure~\ref{fig:lidcex}), where poorly guided decision-making by medical experts can have direct adverse consequences on patients~\cite{BECKER2019108716, brats, jungo2018effect, warfield2004simultaneous, vuadineanu2021analysis, van2016sweet, van2020modeling, boers2022comparing, joskowicz2019inter, watadani2013interobserver}. Given the severity of the involved risks, deep learning models that appropriately deal with aleatoric uncertainty can substantially improve clinical decision-making~\cite{Versteijne2017ConsiderableIV, Joo2019PreoperativeCC, viviers2023segmentation}.

Presenting various plausible image segmentation hypotheses was initially enabled by the use of Monte-Carlo dropout~\cite{kendall2015bayesian}, ensembling methods~\cite{osband2016deep, lakshminarayanan2017simple}, or the use of multiple classification heads~\cite{rupprecht2017learning, ilg2018uncertainty}. However, these methods approximate the weight distribution over the neural networks, i.e., the epistemic uncertainty, rather than the aleatoric uncertainty. In this context, the PU-Net made significant advances by combining the conditional Variational Autoencoder~(VAE)~\cite{kingma2013auto, NIPS2015_8d55a249} and U-Net~\cite{ronneberger2015u}. In this case, a suitable objective is derived from the ELBO of the \textit{conditional} data log-likelihood, which is analogous to the unconditional VAE. The PU-Net is especially interesting in its application, as the low-dimensional latent densities enable interpretability and manipulation of the data, while maintaining a theoretically justified probabilistic framework with fast inference. Nonetheless, several works have recently highlighted the limitations of the PU-Net~\cite{baumgartner2019phiseg, kohl2019hierarchical, kassapis2021calibrated}. Most relevant to this work, which pertains to improving the latent space, is the augmentation of Normalizing Flows to the posterior density to boost expressivity~\cite{valiuddin2021improving, selvan2020uncertainty}. Also, Bhat~\etal~\cite{bhat2023effect} demonstrate changes in model performance subject to density modeling design choices. Beyond this, the latent-space behaviour of the PU-Net has, to the best of our knowledge, not received much attention.
% For example, the Variational Autoencoder (VAE) learns the distribution of the dataset, e.g., the variation of images across classes~\cite{kingma2013auto}. This is achieved by learning latent densities through maximizing a lower bound on the data log-likelihood, known as the Evidence Lower Bound~(ELBO).  As such, the Probabilistic U-Net (PU-Net)~\cite{kohl2018probabilistic} has been proposed. The PU-Net is a widely accepted application for probabilistic segmentation~\cite{liu2022variational,valiuddin2021improving,kohl2019hierarchical,bhat2022generalized,baumgartner2019phiseg,mehrtash2020confidence,selvan2020uncertainty,hu2019supervised}. Nevertheless, the latent space behaviour of the PU-Net has, to the best of our knowledge, not received much attention. 

In this work, we find that the latent space can possess properties that inhibit aleatoric uncertainty quantification. Specifically, the learned latent space variances are significantly sparse. This under-utilization of the latent space can indicate that a substantial amount of information from the posterior is simply ignored. As a consequence, the inhomogeneous latent variance distribution causes the network to be ill-conditioned, resulting in inefficient and unstable gradient descent updates. While this issue is a general concern, its significance within medical settings is particularly pronounced, as underscored by clinical research~\cite{doi1999computer,watadani2013interobserver,rosenkrantz2013comparison,joskowicz2019inter,schaekermann2019understanding,becker2019variability,fu2020retrospective}.

We address this challenge by modifying the ELBO such that it enables the introduction of a new model, the Sinkhorn PU-Net (SPU-Net), whose training objective uses the entropy-regularized Sinkhorn Divergence. These modifications encourage homogeneity in the latent-density singular values, which makes the decoder better conditioned and leads to more effective gradient-descent optimization. As such, the encapsulation of the inter-observer variability is significantly improved.
% As a consequence, the image ambiguity is better encapsulated and the probabilistic segmentation performance is significantly improved. Furthermore, the insights provided in this paper are also applicable to other adaptations of the PU-Net~\cite{kohl2019hierarchical,baumgartner2019phiseg,kassapis2021calibrated}, which can result in additional gains.
To summarize, the contributions of this work are:
\begin{itemize}[noitemsep]
    \item Providing detailed insights into the effect of singular values on the quantification of aleatoric uncertainty with conditional latent density models.
    \item Introducing a new training methodology that improves upon preceding latent variable models for probabilistic segmentation, achieving the best Empirical Wasserstein Distance on every evaluated dataset.
\end{itemize}

The remainder of this paper is structured as follows. First, we present various theoretical frameworks in Section~\ref{sec:theory}. These sections serve as a foundation for our proposed methodology in Section~\ref{sec:methods}. Then, quantitative and qualitative results of the conducted experiments are presented and discussed in Section~\ref{sec:results}. Limitations of our work are mentioned in Section~\ref{sec:limitation} and finally, conclusions are drawn in Section~\ref{sec:conclusion}.
In our experiments, we address this challenge by modifying the training objective. This enables the introduction of a new model, the Sinkhorn PU-Net (SPU-Net), which also uses the entropy-regularized Sinkhorn Divergence. Both the modified objective and entropy regularization encourage homogeneity in the latent-density singular values, which causes the decoder to be better conditioned and leading to more effective gradient-descent optimization. As such, the encapsulation of the inter-observer variability is significantly improved.
% As a consequence, the image ambiguity is better encapsulated and the probabilistic segmentation performance is significantly improved. Furthermore, the insights provided in this paper are also applicable to other adaptations of the PU-Net~\cite{kohl2019hierarchical,baumgartner2019phiseg,kassapis2021calibrated}, which can result in additional gains.
To summarize, the contributions of this work are:
\begin{itemize}[noitemsep]
    \item Providing detailed insights into the effect of singular values on the quantification of aleatoric uncertainty with conditional latent density models.
    \item Introducing a new training methodology that significantly improves (up to 11\%) upon preceding latent variable models for probabilistic segmentation.
\end{itemize}

The remainder of this paper is structured as follows. First, we present various theoretical frameworks in Section~\ref{sec:theory}. These sections serve as a foundation for our proposed methodology in Section~\ref{sec:methods}. Then, quantitative and qualitative results of the conducted experiments are presented and discussed in Section~\ref{sec:results}. Limitations of our work are mentioned in Section~\ref{sec:limitation} and finally, conclusions are drawn in Section~\ref{sec:conclusion}.

\section{Theoretical Background}
\label{sec:theory}
This section first provides an introduction to aleatoric uncertainty quantification in segmentation. Then, the VAE and PU-Net are introduced, where both models maximize a lower-bound on the data log-likelihood. We subsequently mention several challenges that can arise during this optimization process. Additionally, the theory of Optimal Transport will be presented with the intention to enable the use of alternative divergence measures in latent space. In the remainder of this paper, we will use calligraphic letters ($\mathcal{X}$) for sets, capital letters ($X$) for random variables, and lowercase letters ($x$) for specific values. Furthermore, we will denote marginals as $P_X$, probability distributions as $P(X)$, and densities as $p(x)$. Vectors and matrices are distinguished with \textbf{boldface} characters.

\subsection{Aleatoric uncertainty quantification}
Conventional deterministic deep learning models attempt to find function $F:\mathcal{X}\rightarrow\mathcal{Y}$, to map input image $\mathbf{X}\in\mathcal{X}$ to segmentation $\mathbf{Y}\in\mathcal{Y}$. In truth, data is inherently ambiguous and annotators can differ in opinion due to varying experience and expertise. This challenges the assumptions of gold-standard annotations and suggests a probabilistic relationship between the images and segmentation masks. This is most evident in multi-annotated datasets. Probabilistic segmentation aims to capture this relationship through an approximate density $p_{\theta}(\mathbf{y}\mid\mathbf{x})$, parameterized by $\theta$. The simplest and most straightforward approach is interpreting the pixel-wise SoftMax values as parameters of a probability mass function. However, such models cannot sample coherent segmentation masks.

Therefore, a wide spectrum of techniques has been developed to address aleatoric uncertainty quantification in segmentation, each offering its unique strengths and limitations, making the selection of methods contingent upon the specific demands and constraints of downstream tasks. The current state-of-the-art is held by Denoising Diffusion Probabilistic Models (DDPMs)~\cite{zbinden2023stochastic, chen2022analog, rahman2023ambiguous}. Nevertheless, DDPMs are hampered by their time-consuming Markovian sampling procedure, rendering them impractical for the validation of extensive datasets and real-time uncertainty estimation. Naturally, a substantial portion of research on DDPMs is devoted to optimizing and accelerating their inference process~\cite{lee2023minimizing, song2020denoising}.

Models based on Generative Adversarial Networks~\cite{kassapis2021calibrated} are relatively fast, but trade off explicit density estimation and training stability. While Stochastic Segmentation Networks~\cite{monteiro2020stochastic} demonstrate comparable performance to DDPMs, it is essential to recognize their substantial memory requirements, particularly in modeling the covariance matrix, which scales with the dimensionality of the data. 

VAE-based methods~\cite{kohl2018probabilistic, kohl2019hierarchical}, which are the focus of our investigation, exhibit notable advantages. Such models offer a probabilistic framework by using tractable latent representation, ultimately resulting in a semantically interpretable and interpolatable data representation, paired with fast inference.

Another notable approach includes test-time augmentation~\cite{wang2019aleatoric}, which is used to explore the locality of the likelihood function to obtain segmentations. Nonetheless, it has not yielded results equivalent to those achieved by other methods. Auxiliary networks have also been used to model annotation variability~\cite{ji2021learning, Guo2022ModelingAV}, but are beyond the scope of aleatoric uncertainty quantification.

In light of this extensive array of approaches and the inherent strengths and limitations associated with each, it is crucial to focus our efforts on addressing the shortcomings of preferred models for the downstream application, contributing to the advancement of the field. Notably, VAE-based models like the PU-Net~\cite{kohl2018probabilistic} have gained popularity due to the mentioned benefits and the adaptability subject to diverse datasets. This widespread adoption underscores their appeal in the research community~\cite{viviers2023probabilistic, kohl2019hierarchical, valiuddin2021improving, bhat2022generalized, selvan2020uncertainty}.

\subsection{Variational Autoencoders}
Let $\mathbf{X} \in \mathcal{X}$ be an observable variable, taking values in~$\mathbb{R}^{D}$. We define $\mathbf{Z}\in\mathcal{Z}$ as a latent, lower-dimensional representation of $\mathbf{x}$, taking values in $\mathbb{R}^d$. It is assumed that $\mathcal{X}$ possesses a low-dimensional structure in $\mathbb{R}^r$, relative to the high-dimensional ambient space $\mathbb{R}^D$. Thus, it is assumed that $D \gg d \geq r$. In other words, the latent dimensionality $d$ is at least as large as the intrinsic dimensionality $r$ of the data. The VAE~\cite{kingma2013auto} aims to approximate the data distribution through $\mathbf{z}$, by maximizing the Evidence Lower Bound (ELBO) on the data log-likelihood, described by
\begin{subequations}
    \begin{align}
        \log p(\mathbf{x}) &\geq\mathbb{E}_{q_{\bm{\theta}}(\mathbf{z}\mid \mathbf{x})} \left[ \log \frac{p(\mathbf{x},\mathbf{z})}{q_{\bm{\theta}}(\mathbf{z}\mid \mathbf{x})} \right] \notag \\
        &=\mathbb{E}_{q_{\bm{\theta}}(\mathbf{z}\mid \mathbf{x})}\left[ \log p_{\bm{\phi}} (\mathbf{x}\mid \mathbf{z})\right]  -\mathrm{KL}\left[q_{\bm{\theta}} (\mathbf{z}\mid \mathbf{x}) \,\mid\mid\, p(\mathbf{z})\right], \tag{1} \label{eq:elbo}
    \end{align}
\end{subequations}
with tractable encoder $q_{\bm{\theta}} (\mathbf{z}\mid \mathbf{x})$ (commonly known as the posterior) and decoder $p_{\bm{\phi}} (\mathbf{x}\mid \mathbf{z})$ densities, amortized with parameters $\bm{\theta}$ and $\bm{\phi}$, respectively. Furthermore, $p(\mathbf{z})$ is a fixed latent density and is commonly referred to as the prior. Under the mean-field approximation, the posterior factorizes across latent dimensions, and the densities are modeled as axis-aligned Normals. Furthermore, to enable prediction from test images, optimization is performed through amortization with neural networks. Nevertheless, due to these approximations, this construction can be sub-optimal, reduce the effectiveness of the VAE and void the guarantee that high ELBO values indicate accurate inference~\cite{zhao2017infovae,cremer2018inference}.

\subsection{The Probabilistic U-Net}\label{sec:probunet}
Similar to the VAE, consider input image and ground-truth segmentation masks $\mathbf{X},\mathbf{Y}\in \mathbb{R}^D$. Then, the ELBO of the \textit{conditional} data log-likelihood can be written as
\begin{subequations}\label{eq:elbo-punet}
    \begin{align}
        \log p(\mathbf{y}\mid\mathbf{x}) &\geq\mathbb{E}_{q_{\bm{\theta}}(\mathbf{z}\mid \mathbf{y},\mathbf{x})} \left[ \log \frac{p(\mathbf{y},\mathbf{z}\mid\mathbf{x})}{q_{\bm{\theta}}(\mathbf{z}\mid \mathbf{y},\mathbf{x})} \right] \notag \\
        &=\mathbb{E}_{q_{\bm{\theta}}(\mathbf{z}\mid \mathbf{y},\mathbf{x})}\left[ \log p_{\bm{\phi}} (\mathbf{y}\mid \mathbf{x}, \mathbf{z})\right] \notag\\
        &\qquad-\mathrm{KL}\left[q_{\bm{\theta}} (\mathbf{z}\mid \mathbf{y}, \mathbf{x}) \,\mid\mid\, p_{\bm{\psi}}(\mathbf{z}\mid\mathbf{x})\right], \tag{2}
    \end{align}
\end{subequations}
and the predictive distribution from test image $\mathbf{x}^*$ as
\begin{equation}
    \begin{aligned}
    p(\mathbf{y}\mid\mathbf{x}^*):=\int_{\mathbb{R}^d}p_{\bm{\phi}}(\mathbf{y}\mid\mathbf{z},\mathbf{x}^*)p_{\bm{\psi}}(\mathbf{z}\mid\mathbf{x}^*)d\mathbf{z}.
    \end{aligned}
\end{equation}

The success of the PU-Net for probabilistic segmentation can be accredited to several additional design choices in the implementation of the encoding-decoding structure. Firstly, an encoding~$\mathbf{z}$, is inserted pixel-wise at the final stages of a U-Net, followed by feature-combining 1$\times$1 convolutions only. As a result, this significantly alters the $r$-dimensional manifold that $\mathbf{z}$ attempts to learn. Namely, it learns the segmentation variability within a single image rather than features of the segmentation itself. Therefore, much smaller values of $d$ are feasible. Another crucial difference in the PU-Net formulation w.r.t. that of the VAE is the conditioning of the prior $p_{\bm{\psi}}(\mathbf{z}\mid\mathbf{x})$, which is not fixed but rather learned. Although it has been argued by Zheng~\etal~\cite{zheng2022learning} that learning the prior density is theoretically equivalent to using a non-trainable prior, their respective optimization trajectories can differ substantially such that the former method is preferred~\cite{kohl2018probabilistic, kohl2019hierarchical, baumgartner2019phiseg}. 

\subsection{Challenges with latent variable models}\label{sec:challenges}
\subsubsection{Information preference}
A common hurdle when training the VAE is the phenomenon of ignored latent codes~\cite{chen2016variational, zhao2017infovae}. While ignored latent variables can potentially produce realistic looking samples, they completely undermine the model's inference capabilities relevant for downstream tasks. This problem was initially recognized when strong networks such as Recurrent Neural Networks (RNNs) were used as decoders~\cite{bowman2015generating, serban2017hierarchical, fraccaro2016sequential}. A common hypothesis argues that early on during training, the latent codes $\mathbf{z}$ carry negligible information on $\mathbf{x}$. Therefore, the prior regularization term in the ELBO objective can be easily minimized. Chen~\etal~\cite{chen2016variational} describe that this problem is not solely related to optimization, but can generally occur subject to intractable true posteriors and powerful decoders. Zhao~\etal~\cite{zhao2017infovae} also demonstrate that the ELBO can reach global optimum without utilizing the latent codes. The PU-Net, an extension of the ELBO to the conditional case, can therefore suffer from similar issues such that it defaults to non-informative latent variables.

\subsubsection{Learnable priors}
Instead of fixing the prior, learning this density has shown commendable results. Nonetheless, such modeling choice can exacerbate collapse to degenerate solutions with uninformative latent variables. Consider the KL-divergence between two $d$-dimensional axis-aligned Normals $\mathcal{N}_1$ and $\mathcal{N}_2$ with identical means 
\begin{equation}\label{eq:collapsenormal}
    \begin{aligned}
        2\cdot\operatorname{KL}\left[\,\mathcal{N}_1 \,||\, \mathcal{N}_2\,\right]=\sum_i \frac{\sigma^i_1}{\sigma^i_2} + \ln \frac{ \sigma^i_2}{\sigma^i_1} - d.
    \end{aligned}
\end{equation}
Minimizing this expression entails finding $\sigma^i_1=\sigma^i_2$ for all values of $i$. At the same time, the singular values are progressively minimized~\cite{zheng2022learning} and the variance vectors $\bm{\sigma}_1$ and $\bm{\sigma}_2$ are not constrained in any fashion. This means the variances can be arbitrarily distributed, including low-entropy solutions (i.e. sparse) where the latent space is completely ignored except for some dimension that governs the required stochasticity to satisfy the reconstruction cost in Equation~(\ref{eq:elbo-punet}).

\subsubsection{Decoder sensitivity}\label{subsubseq:sensitivity}
A sparse latent representation can cause the decoder to become extremely sensitive to the latent-space sample localization. The PU-Net decoder is dependent on the function $\mathbf{f}_{\bm{{\phi}}}(\mathbf{x},\mathbf{z})$, where $p_{\bm{\phi}} (\mathbf{y}\mid \mathbf{x}, \mathbf{z})=\delta(\mathbf{y}-\mathbf{f}_{\bm{{\phi}}}(\mathbf{x},\mathbf{z}))$, which intermediately involves latent variable $\mathbf{z}$ to reconstruct a plausible segmentation hypothesis. The latent variable $\mathbf{z}$ is controlled by the mean $\bm{\mu}$ and singular values $\bm{\sigma}$ of the underlying axis-aligned Normal density. If the input $\mathbf{x}$ is considered to be fixed, then the relative condition number, $\zeta_i$, of $\mathbf{f}_{\bm{{\phi}}}(\mathbf{x},\mathbf{z})$ can be interpreted as the sensitivity of its output w.r.t. its varying input $z_i$, corresponding to the $i$-th latent dimension, which can be expressed as 
% \begin{equation}
%     \lim_{\epsilon\rightarrow 0}\sup_{||\delta \mathbf{z}||\leq \epsilon} \frac{||\delta \mathbf{h}(\mathbf{g}(\mathbf{x}),\mathbf{z}))||}{||\delta \mathbf{z}||}.
% \end{equation}
\begin{equation}
    \zeta_i = \lim_{\epsilon\rightarrow 0}\sup_{||\delta z_i||\leq \epsilon} \frac{||\delta \mathbf{f}_{\bm{{\phi}}}(\mathbf{x}, z_i)|| \;/\; ||\mathbf{f}_{\bm{{\phi}}}(\mathbf{x},\mathbf{z})||}{||\delta z_i || \;/\; ||\mathbf{z}||}.
\end{equation} 
For convenient notation, only the varying $z_i$ is included in $\delta \mathbf{f}_{\bm{{\phi}}}(\mathbf{x}, z_i)$ rather than the complete vector $\mathbf{z}$. The condition number of a function has direct consequences to the numerical stability of its gradient updates. Suppose $\sigma_1 \gg \sigma_2$ and $\mu_1\approx\mu_2$, then $\zeta_2 \gg \zeta_1$, and the function is said to be ill-conditioned. To illustrate with a simplification, consider the function
\begin{equation}
    f(z_1, z_2) = w_1z_1 + w_2z_2
\end{equation}
with weights \( w_1 \) and \( w_2 \), and latent variables $z_1$ and $z_2$. The subsequent gradients are
\begin{equation}
    \nabla f = \left[ \frac{\partial f}{\partial w_1}, \frac{\partial f}{\partial w_2} \right]^T = \left[ z_1, z_2 \right]^T.
\end{equation} 
It can be seen that the gradient updates are governed by the magnitudes of $z_1$ and $z_2$. Hence, the case $\sigma_1 \gg \sigma_2$ (the underlying variances of $z$) will skew the optimization landscape, where the gradients w.r.t. $w_1$ will dominate, leading to inefficient and unstable gradient descent updates. When considering that the dimensionality is usually higher than $d=2$, the search towards an optimal minimum will rapidly suffer from irregular condition numbers~\cite{alger2019data}.

In conventional convex optimization, Newton's method is used to consider curvature of the optimization landscape, appropriately dealing with inefficient gradient updates due to skewness. In deep neural networks, normalization layers have been used to smooth optimization and have led to considerable improvements~\cite{ioffe2015batch, salimans2016weight, ba2016layer, wu2018group, qiao2019micro}. In the case of latent variable modeling, this pertains to encouraging latent variances that are approximately of the same magnitude.
\subsubsection{Augmentation with Normalizing Flows}
Furthermore, past literature has made mention of enhancing the posterior density by augmentation of Normalizing Flows~\cite{rezende2015variational}, which was proposed to enhance the expressivity and complexity of the posterior distribution to reduce the amortization gap~\cite{cremer2018inference}. This has been shown to improve the PU-Net~\cite{valiuddin2021improving, selvan2020uncertainty}. Nevertheless, it has been argued in previous works that the mean-field Gaussian assumption in the VAE is not necessarily the cause for the failure to learn the ground-truth manifold~\cite{dai2019diagnosing}. Thus, further exploration on the effects of an NF-augmented posterior is required.
% We hypothesize that this is related to the latent dimension usage of the posterior density, where the introduced technique regularizes the relative condition numbers of the decoder. In the context of the gradient descent, we can consider augmenting with NFs as a form of preconditioning, where it smoothens optimization by an appropriate transformation of the latent samples. 
\subsection{Optimal Transport}\label{subsec:OT}
Let $\mathcal{Y}$ and $\mathcal{\hat{Y}}$ be the two separable metric spaces. For the sake of clarity, the reader can assume that these sets contain the ground-truth and model predictions, respectively. We adopt the Monge-Kantorovich formulation~\cite{villani2008optimal} for the OT problem, by specifying
\begin{subequations}\label{eq:monge}
\begin{align}
    W(\mu, \hat{\mu}&):= \notag \\ 
    &\operatorname{inf}\left\{\int_{\mathcal{Y}\times\mathcal{\hat{Y}}} c(\mathbf{y},\hat{\mathbf{y}})\,d\gamma(\mathbf{y},\hat{\mathbf{y}})\bigg| \gamma \in \Gamma(\mu,\hat{\mu})\right\} \tag{8},
    \end{align}
\end{subequations}
where $\gamma \in \Gamma(\mu,\hat{\mu})$ denotes a coupling in the tight collection of all probability measures on $\mathcal{Y} \times \mathcal{\hat{Y}}$ with  marginals $\mu$ and $\hat{\mu}$, respectively. The function $c(\mathbf{y},\hat{\mathbf{y}})$: $\mathcal{Y} \times \mathcal{\hat{Y}}\rightarrow\mathbb{R}_+$ denotes any lower semi-continuous measurable cost function. Equation~(\ref{eq:monge}) is also commonly referred to as the Wasserstein distance. Furthermore, the usual context of this formulation is in finding the lowest cost of moving samples from the probability measures in $\mathcal{Y}$ to the measures in $\mathcal{\hat{Y}}$. In the case of probabilistic segmentation, the aim is to learn the ground-truth distribution $P_{Y\mid X}$, by matching it with the model distribution $P_{\hat{Y} \mid X}$. The Wasserstein distance is in practice a tedious calculation. As a solution, it has been proposed to introduce entropic regularization~\cite{wilson1968use}. This is achieved by using the entropy of the couplings $\gamma$ as a regularizing function, which is specified by
\begin{equation}
\begin{aligned}
\tilde{S}_{\epsilon}(\mu, \hat{\mu}&):= \\ 
&\operatorname{inf} \bigg\{ \int_{\mathcal{Y}\times\mathcal{\hat{Y}}} s(\mathbf{y},\hat{\mathbf{y}})\,d\gamma(\mathbf{y},\hat{\mathbf{y}})\bigg | \gamma \in \Gamma(\mu,\hat{\mu}) \bigg\},
\end{aligned}
\end{equation}
where
\begin{equation}
    s(\mathbf{y},\mathbf{\hat{y}}) = c(\mathbf{y},\mathbf{\hat{y}}) + \epsilon \operatorname{log}\frac{d\gamma(\mathbf{y},\hat{\mathbf{y}})}{d\mu(\mathbf{y}) d\hat{\mu}(\mathbf{\hat{y}})}.
\end{equation}
As mentioned by Cuturi~\cite{cuturi2013sinkhorn}, the entropy term can be expanded to $\operatorname{log}(d\gamma(\mathbf{y},\hat{\mathbf{y}}))-\operatorname{log}(d\mu(\mathbf{y}))-\operatorname{log}(d\hat{\mu}(\mathbf{\hat{y}}))$. Cuturi~\cite{cuturi2013sinkhorn} shows that this entropic regularization allows optimization over a Lagrangian dual for faster computation with the iterative Sinkhorn matrix scaling algorithm. Additionally, the entropic bias is removed from the OT problem to obtain the Sinkhorn Divergence, specified as
\begin{equation}\label{eq:sinkhorndiver}
    S_{\epsilon}(\mu, \hat{\mu}) = \tilde{S}_{\epsilon}(\mu, \hat{\mu}) - \frac{1}{2}\left(\tilde{S}_{\epsilon}(\mu, \mu) + \tilde{S}_{\epsilon}(\hat{\mu}, \hat{\mu}) \right). 
\end{equation}
The Sinkhorn Divergence interpolates between $W$ when $\epsilon\rightarrow~0$ with $\mathcal{O}(\epsilon\operatorname{log}(\frac{1}{\epsilon}))$ deviation, and Maximum Mean Discrepancy when $\epsilon\rightarrow\infty$, which favours dimension-independent sample complexity~\cite{genevay2019sample}. A viable option is to approximate the Sinkhorn Divergence via sampling with weights $a,b \in \mathbb{R}_+$. To increase efficiency and speed, Kosowsky and Yuille~\cite{annealing} introduce \textit{$\epsilon$-scaling} or \textit{simulated annealing} to the Sinkhorn algorithm, which implies letting $\epsilon$ decrease across iterations.

The entropic regularization can be understood as constraining the joint probability distribution to have \textit{sufficient entropy}, or contain small enough \textit{mutual information} with respect to $d\mu$ and $d\hat{\mu}$. Assuming normally distributed joint density $\mathcal{N}$ with covariance $\Sigma$ for the sake of analysis, the mutual information in the case of identical means becomes
\begin{equation}
    I(\mathbf{z}_1,\mathbf{z}_2)= \mathrm{KL}[\mathcal{N} \,||\, \mathcal{N}_1 \mathcal{N}_2] = \frac{1}{2}\log\frac{\det \Sigma_1 \det \Sigma_2}{\det \Sigma},
\end{equation}
with $\Sigma$ being the covariance of the joint probability density. This can be considered as a multivariate correlation measure, as the expression in the case of $d=1$ reduces to $\log \sqrt{\frac{1}{1-r^2}}$, where $r$ is the correlation coefficient. Thus, strong correlation between source and destination in the transport plan is penalized by entropic regularization, preventing collapse to deterministic, low-entropy couplings.

Various works have investigated constraining the latent densities by means of Optimal Transport~\cite{bousquet2017optimal}. Tolstikhin~\etal~\cite{tolstikhin2019wasserstein} introduce Wasserstein Autoencoders~(WAEs), which softly constrain the latent densities with a Wasserstein penalty term, a metric that emerged from OT theory~\cite{vaserstein1969markov}. The authors demonstrate that the WAEs exhibit better sample quality while maintaining the favourable properties of the VAE. Similarly, Patrini~\etal~\cite{patrini2019sinkhorn} introduce the Sinkhorn Autoencoder~(SAE), in which the Sinkhorn algorithm~\cite{cuturi2013sinkhorn} -- an approximation of the Wasserstein distance -- is leveraged as a latent divergence measure. The aforementioned research has been conducted in the setting of a VAE with an unconditional ELBO objective. The PU-Net architecture, on the other hand, is an extension of the conditional VAE (cVAE), which employs the conditional ELBO with a learnable prior density and alters the nature of the training objective.
% For this, it is beneficial to consider the problem from the perspective of Optimal Transport, rather than just maximization of the data log-likelihood, such that utilization of alternative latent-space constraints are justified. This will change the manner in which the aleatoric uncertainty is encapsulated by the latent densities.VAE-based methods~\cite{kohl2018probabilistic, kohl2019hierarchical}, which are the focus of our investigation, exhibit notable advantages. Such models offer a probabilistic framework by using tractable latent representation, ultimately resulting in a semantically interpretable and interpolatable data representation, paired with fast inference.

Another notable approach includes test-time augmentation~\cite{wang2019aleatoric}, which is used to explore the locality of the likelihood function to obtain segmentations. Nonetheless, it has not yielded results equivalent to those achieved by other methods. Auxiliary networks have also been used to model annotation variability~\cite{ji2021learning, Guo2022ModelingAV}, but are beyond the scope of aleatoric uncertainty quantification.

In light of this extensive array of approaches and the inherent strengths and limitations associated with each, it is crucial to focus our efforts on addressing the shortcomings of preferred models for the downstream application, contributing to the advancement of the field. Notably, models VAE-based models like the PU-Net~\cite{kohl2018probabilistic} have gained popularity due to the mentioned benefits and the adaptability subject to diverse datasets. This widespread adoption underscores their appeal in the research community~\cite{viviers2023probabilistic, kohl2019hierarchical, valiuddin2021improving, bhat2022generalized, selvan2020uncertainty}.

\subsection{Variational Autoencoders}
Let $\mathbf{X} \in \mathcal{X}$ be an observable variable, taking values in~$\mathbb{R}^{D}$. We define $\mathbf{Z}\in\mathcal{Z}$ as a latent, lower-dimensional representation of $\mathbf{x}$, taking values in $\mathbb{R}^d$. It is assumed that $\mathcal{X}$ possesses a low-dimensional structure in $\mathbb{R}^r$, relative to the high-dimensional ambient space $\mathbb{R}^D$. Thus, it is assumed that $D \gg d \geq r$. In other words, the latent dimensionality $d$ is greater than the intrinsic dimensionality $r$ of the data. The VAE~\cite{kingma2013auto} aims to approximate the data distribution through $\mathbf{z}$, by maximizing the Evidence Lower Bound (ELBO) on the data log-likelihood, described by
\begin{subequations}
    \begin{align}
        \log p(\mathbf{x}) &\geq\mathbb{E}_{q_{\bm{\theta}}(\mathbf{z}\mid \mathbf{x})} \left[ \log \frac{p(\mathbf{x},\mathbf{z})}{q_{\bm{\theta}}(\mathbf{z}\mid \mathbf{x})} \right] \notag \\
        &\geq\mathbb{E}_{q_{\bm{\theta}}(\mathbf{z}\mid \mathbf{x})}\left[ \log p_{\bm{\phi}} (\mathbf{x}\mid \mathbf{z})\right]  -\mathrm{KL}\left[q_{\bm{\theta}} (\mathbf{z}\mid \mathbf{x}) \,\mid\mid\, p(\mathbf{z})\right], \tag{1} \label{eq:elbo}
    \end{align}
\end{subequations}
with tractable encoder $q_{\bm{\theta}} (\mathbf{z}\mid \mathbf{x})$ (commonly known as the posterior) and decoder $p_{\bm{\phi}} (\mathbf{x}\mid \mathbf{z})$ densities, amortized with parameters $\bm{\theta}$ and $\bm{\phi}$, respectively. Furthermore, $p(\mathbf{z})$ is a fixed latent density and is commonly referred to as the prior. As a consequence of the mean-field approximation, the densities are modeled with axis-aligned Normal densities. Furthermore, to enable prediction from test images, optimization is performed through amortization with neural networks. Nevertheless, due to these approximations, this construction can be sub-optimal, reduce the effectiveness of the VAE and void the guarantee that high ELBO values indicate accurate inference~\cite{zhao2017infovae,cremer2018inference}.

\subsection{The Probabilistic U-Net}\label{sec:probunet}
Similar to the VAE, consider input image and ground-truth segmentation masks $\mathbf{X},\mathbf{Y}\in \mathbb{R}^D$. Then, the ELBO of the \textit{conditional} data log-likelihood can be written as
\begin{subequations}\label{eq:elbo-punet}
    \begin{align}
        \log p(\mathbf{y}\mid\mathbf{x}) &\geq\mathbb{E}_{q_{\bm{\theta}}(\mathbf{z}\mid \mathbf{x},\mathbf{y})} \left[ \log \frac{p(\mathbf{y},\mathbf{z}\mid\mathbf{x})}{q_{\bm{\theta}}(\mathbf{z}\mid \mathbf{y},\mathbf{x})} \right] \notag \\
        &\geq\mathbb{E}_{q_{\bm{\theta}}(\mathbf{z}\mid \mathbf{x},\mathbf{y})}\left[ \log p_{\bm{\phi}} (\mathbf{y}\mid \mathbf{x}, \mathbf{z})\right] \notag\\
        &\qquad-\mathrm{KL}\left[q_{\bm{\theta}} (\mathbf{z}\mid \mathbf{y}, \mathbf{x}) \,\mid\mid\, p_{\bm{\psi}}(\mathbf{z}\mid\mathbf{x})\right], \tag{2}
    \end{align}
\end{subequations}
and the predictive distribution from test image $\mathbf{x}^*$ as
\begin{equation}
    \begin{aligned}
    p(\mathbf{y}\mid\mathbf{x}^*):=\int_{\mathbb{R}^d}p_{\bm{\phi}}(\mathbf{y}\mid\mathbf{z},\mathbf{x}^*)p_{\bm{\psi}}(\mathbf{z}\mid\mathbf{x}^*)d\mathbf{z}.
    \end{aligned}
\end{equation}

The success of the PU-Net for probabilistic segmentation can be accredited to several additional design choices in the implementation of the encoding-decoding structure. Firstly, an encoding~$\mathbf{z}$, is inserted pixel-wise at the final stages of a U-Net, followed by feature-combining 1$\times$1 convolutions only. As a result, this significantly alters the $r$-dimensional manifold that $\mathbf{z}$ attempts to learn. Namely, it learns the segmentation variability within a single image rather than features of the segmentation itself. Therefore, much smaller values of $d$ are feasible. Another crucial difference in the PU-Net formulation w.r.t. to that of the VAE is the conditioning of the prior $p_{\bm{\psi}}(\mathbf{z}\mid\mathbf{x})$, which is not fixed but rather learned. Although it has been argued by Zheng~\etal~\cite{zheng2022learning} that learning the prior density is theoretically equivalent to using a non-trainable prior, their respective optimization trajectories can differ substantially such that the former method is preferred~\cite{kohl2018probabilistic, kohl2019hierarchical, baumgartner2019phiseg}. 

\subsection{Challenges with latent variable models}\label{sec:challenges}
\subsubsection{Information preference}
A common hurdle when training the VAE is the phenomena of ignored latent codes~\cite{chen2016variational, zhao2017infovae}. While ignored latent variables can potentially produce realistic looking samples, it completely undermines its inference capabilities relevant for downstream tasks. This problem was initially recognized when strong networks such as Recurrent Neural Networks (RNNs) were used as decoders~\cite{bowman2015generating, serban2017hierarchical, fraccaro2016sequential}. A common hypothesis argues that early on during training, the latent codes $\mathbf{z}$ carry negligent information on $\mathbf{x}$. Therefore, the prior regularization term in the ELBO the objective can be easily minimized. Chen~\etal~\cite{chen2016variational} describe that this problem is not solely related to optimization, but can generally occur subject to intractable true posteriors and powerful decoders. Zhao~\etal~\cite{zhao2017infovae} also demonstrate that the ELBO can reach global optimum without utilizing the latent codes. The PU-Net, an extension of the ELBO to the conditional case, can therefore suffer from similar issues such that it defaults to non-informative latent variables.

\subsubsection{Learnable priors}
Instead of fixing the prior, learning this density has shown commendable results. Nonetheless, such modeling choice can exacerbate collapse to degenerate solutions with uninformative latent variables. Consider the KL-divergence between two $d$-dimensional axis-aligned Normals $\mathcal{N}_1$ and $\mathcal{N}_2$ with identical means 
\begin{equation}\label{eq:collapsenormal}
    \begin{aligned}
        2\cdot\operatorname{KL}\left[\,\mathcal{N}_1 \,||\, \mathcal{N}_2\,\right]=\sum_i \frac{\sigma^i_2}{\sigma^i_1} + \ln \frac{ \sigma^i_1}{\sigma^i_2} - d.
    \end{aligned}
\end{equation}
Minimizing this expression entails finding $\sigma^i_1=\sigma^i_2$ for all values of $i$. At the same time, the singular values are progressively minimized~\cite{zheng2022learning} and the variance vectors $\bm{\sigma}_1$ and $\bm{\sigma}_2$ are not constrained in any fashion. This means the variances can be arbitrarily distributed, including low-entropy solutions (i.e. sparse) where the latent space is completely ignored except for some dimension that governs the required stochasticity to satisfy the reconstruction cost in Equation~(\ref{eq:elbo-punet}).

\subsubsection{Decoder sensitivity}\label{subsubseq:sensitivity}
A sparse latent representation can cause the decoder to become extremely sensitive to the latent-space sample localization. The PU-Net decoder is dependent on the function $\mathbf{f}_{\bm{{\phi}}}(\mathbf{x},\mathbf{z})$, where $p_{\bm{\phi}} (\mathbf{y}\mid \mathbf{x}, \mathbf{z})=\delta(\mathbf{y}-\mathbf{f}_{\bm{{\phi}}}(\mathbf{x},\mathbf{z}))$, which intermediately involves latent variable $\mathbf{z}$ to reconstruct a plausible segmentation hypothesis. The latent variable $\mathbf{z}$ is controlled by the mean $\bm{\mu}$ and singular values $\bm{\sigma}$ of the underlying axis-aligned Normal density. If the input $\mathbf{x}$ is considered to be fixed, then the relative condition number, $\zeta_i$, of $\mathbf{f}_{\bm{{\phi}}}(\mathbf{x},\mathbf{z})$ can be interpreted as the sensitivity of its output w.r.t. its varying input $z_i$, corresponding to the $i$-th latent dimension, which can be expressed as 
% \begin{equation}
%     \lim_{\epsilon\rightarrow 0}\sup_{||\delta \mathbf{z}||\leq \epsilon} \frac{||\delta \mathbf{h}(\mathbf{g}(\mathbf{x}),\mathbf{z}))||}{||\delta \mathbf{z}||}.
% \end{equation}
\begin{equation}
    \zeta_i = \lim_{\epsilon\rightarrow 0}\sup_{||\delta z_i||\leq \epsilon} \frac{||\delta \mathbf{f}_{\bm{{\phi}}}(\mathbf{x}, z_i)|| \;/\; ||\mathbf{f}_{\bm{{\phi}}}(\mathbf{x},\mathbf{z})||}{||\delta z_i || \;/\; ||\mathbf{z}||}.
\end{equation} 
For convenient notation, only the varying $z_i$ is included in $\delta \mathbf{f}_{\bm{{\phi}}}(\mathbf{x}, z_i)$ rather than the complete vector $\mathbf{z}$. The condition number of a function has direct consequences to the numerical stability of its gradient updates. Suppose $\sigma_1 \gg \sigma_2$ and $\mu_1\approx\mu_2$, then $\zeta_2 \gg \zeta_1$, and the function is said to be ill-conditioned. To illustrate with a simplification, consider function
\begin{equation}
    f(z_1, z_2) = w_1z_1 + w_2z_2
\end{equation}
with weights \( w_1 \) and \( w_2 \), and latent variables $z_1$ and $z_2$. The subsequent gradients are
\begin{equation}
    \nabla f = \left[ \frac{\partial f}{\partial w_1}, \frac{\partial f}{\partial w_2} \right]^T = \left[ z_1, z_2 \right]^T.
\end{equation} 
It can be seen that the gradients updates are governed by the magnitudes of $z_1$ and $z_2$. Hence, the case $\sigma_1 \gg \sigma_2$ (the underlying variances of $z$) will skew the optimization landscape, where the gradients w.r.t. $z_2$ will dominate, leading to inefficient and unstable gradient descent updates. When considering that the dimensionality is usually higher than $d=2$, the search towards an optimal minimum will rapidly suffer from irregular condition numbers~\cite{alger2019data}.

In conventional convex optimization, the Newton's method is used to consider curvature of the optimization landscape, appropriately dealing with inefficient gradient updates due to skewness. In complex and deep neural networks, normalization layers have been used in neural networks to smoothen optimization and have led to considerable improvements~\cite{ioffe2015batch, salimans2016weight, ba2016layer, wu2018group, qiao2019micro}. In the case of latent variable modeling, this pertains stimulating latent variances which are approximately of the same magnitude.
\subsubsection{Augmentation with Normalizing Flows}
Furthermore, past literature has made mention of enhancing the posterior density by augmentation of Normalizing Flows~\cite{rezende2015variational}, which was proposed to enhance the expressivity and complexity of the posterior distribution to reduce the amortization gap~\cite{cremer2018inference}. This has shown to improve the PU-Net~\cite{valiuddin2021improving, selvan2020uncertainty}. Nevertheless, it has been argued in previous works that the mean-field Gaussian assumption in the VAE is not necessarily the cause for the failure to learn the ground-truth manifold~\cite{dai2019diagnosing}. Thus, further exploration on the effects on a NF-augmented posterior is required.
% We hypothesize that this is related to the latent dimension usage of the posterior density, where the introduced technique regularizes the relative condition numbers of the decoder. In the context of the gradient descent, we can consider augmenting with NFs as a form of preconditioning, where it smoothens optimization by an appropriate transformation of the latent samples. 
\subsection{Optimal Transport}\label{subsec:OT}
Let $\mathcal{Y}$ and $\mathcal{\hat{Y}}$ be the two separable metric spaces. For the sake of clarity, the reader can assume that these sets contain the ground-truth and model predictions, respectively. We adopt the Monge-Kantorovich formulation~\cite{villani2008optimal} for the OT problem, by specifying
\begin{subequations}\label{eq:monge}
\begin{align}
    W(\mu, \hat{\mu}&):= \notag \\ 
    &\operatorname{inf}\left\{\int_{\mathcal{Y}\times\mathcal{\hat{Y}}} c(\mathbf{y},\hat{\mathbf{y}})\,d\gamma(\mathbf{y},\hat{\mathbf{y}})\bigg| \gamma \in \Gamma(\mu,\hat{\mu})\right\} \tag{8},
    \end{align}
\end{subequations}
where $\gamma \in \Gamma(\mu,\hat{\mu})$ denotes a coupling in the tight collection of all probability measures on $\mathcal{Y} \times \mathcal{\hat{Y}}$ with  marginals $\mu$ and $\hat{\mu}$, respectively. The function $c(\mathbf{y},\hat{\mathbf{y}})$: $\mathcal{Y} \times \mathcal{\hat{Y}}\rightarrow\mathbb{R}_+$ denotes any lower semi-continuous measurable cost function. Equation~(\ref{eq:monge}) is also commonly referred to as the Wasserstein distance. Furthermore, the usual context of this formulation is in finding the lowest cost of moving samples from the probability measures in $\mathcal{Y}$ to the measures in $\mathcal{\hat{Y}}$. In the case of probabilistic segmentation, the aim is to learn the ground-truth distribution $P_{Y\mid X}$, by matching it with the model distribution $P_{\hat{Y} \mid X}$. The Wasserstein distance is in practice a tedious calculation. As a solution, it has been proposed to introduce entropic regularization~\cite{wilson1968use}. This is achieved by using the entropy of the couplings $\gamma$ as a regularizing function, which is specified by
\begin{equation}
\begin{aligned}
\tilde{S}_{\epsilon}(\mu, \hat{\mu}&):= \\ 
&\operatorname{inf} \bigg\{ \int_{\mathcal{Y}\times\mathcal{\hat{Y}}} s(\mathbf{y},\mathbf{\hat{y})}d\gamma(\mathbf{y},\hat{\mathbf{y}})\bigg | \gamma \in \Gamma(\mu,\hat{\mu}) \bigg\},
\end{aligned}
\end{equation}
where
\begin{equation}
    s(\mathbf{y},\mathbf{\hat{y}}) = d(\mathbf{y},\mathbf{\hat{y}}) + \epsilon \operatorname{log}\frac{d\gamma(\mathbf{y},\hat{\mathbf{y}})}{d\mu(\mathbf{y}) d\hat{\mu}(\mathbf{\hat{y}})}.
\end{equation}
As mentioned by Cuturi~\etal~\cite{cuturi2013sinkhorn}, the entropy term can be expanded to $\operatorname{log}(d\gamma(\mathbf{y},\hat{\mathbf{y}}))-\operatorname{log}(d\mu(\mathbf{y}))-\operatorname{log}(d\hat{\mu}(\mathbf{\hat{y}}))$. Cuturi~\etal~\cite{cuturi2013sinkhorn} show that this entropic regularization allows optimization over a Lagrangian dual for faster computation with the iterative Sinkhorn matrix scaling algorithm. Additionally, the entropic bias is removed from the OT problem to obtain the Sinkhorn Divergence, specified as
\begin{equation}\label{eq:sinkhorndiver}
    S_{\epsilon}(\mu, \hat{\mu}) = \tilde{S}_{\epsilon}(\mu, \hat{\mu}) - \frac{1}{2}\left(\tilde{S}_{\epsilon}(\mu, \mu) + \tilde{S}_{\epsilon}(\hat{\mu}, \hat{\mu}) \right). 
\end{equation}
The Sinkhorn Divergence interpolates between $W_p$ when $\epsilon\rightarrow~0$ with $\mathcal{O}(\epsilon\operatorname{log}(\frac{1}{\epsilon}))$ deviation, and Maximum Mean Discrepancy when $\epsilon\rightarrow\infty$, which favours dimension-independent sample complexity~\cite{genevay2019sample}. A viable option is to approximate the Sinkhorn Divergence via sampling with weights $\alpha,\beta \in \mathbb{R}_+$. To increase efficiency and speed, Kosowsky and Yuille~\cite{annealing} introduce \textit{$\epsilon$-scaling} or \textit{simulated annealing} to the Sinkhorn algorithm, which implies letting $\epsilon$ decrease across iterations.

The entropic regularization can be understood as constraining the joint probability distribution to have \textit{sufficient entropy}, or contain small enough \textit{mutual information} with respect to $d\mu$ and $d\hat{\mu}$. Assuming normally distributed joint density $\mathcal{N}$ with covariance $\Sigma$ for the sake of analysis, the mutual information in the case of identical means becomes
\begin{equation}
    I(\mathbf{z}_1,\mathbf{z}_2)= D_{KL}[\mathcal{N} \,||\, \mathcal{N}_1 \mathcal{N}_2] = \frac{\det \Sigma_1 \det \Sigma_2}{\det \Sigma},
\end{equation}
with $\Sigma$ being the convariance of the joint probability density. This can be considered as a multivariate correlation measure, as the expression in the case of $d=1$ reduces to $\log \sqrt{\frac{1}{1-r^2}}$, where $r$ is the correlation coefficient. Thus, strong correlation between the posterior and prior latent variables will be penalized by entropic regularization, preventing collapse to low-entropy solutions and as mentioned in Section~\ref{subsubseq:sensitivity}, will improve gradient descent updates.

Various works have investigated constraining the latent densities by means of Optimal Transport~\cite{bousquet2017optimal}. Tolstikhin~\etal~\cite{tolstikhin2019wasserstein} introduce Wasserstein Autoencoders~(WAEs), which softly constrain the latent densities with a Wasserstein penalty term, a metric that emerged from OT theory~\cite{vaserstein1969markov}. The authors demonstrate that the WAEs exhibit better sample quality while maintaining the favourable properties of the VAE. Similarly, Patrini~\etal~\cite{patrini2019sinkhorn} introduce the Sinkhorn Autoencoder~(SAE), in which the Sinkhorn algorithm~\cite{cuturi2013sinkhorn} -- an approximation of the Wasserstein distance -- is leveraged as a latent divergence measure. The aforementioned research has been conducted in the setting of a VAE with an unconditional ELBO objective. The PU-Net architecture, on the other hand, is an extension of the cVAE, which employs the conditional ELBO with a learnable prior density and alters the nature of the training objective.
% For this, it is beneficial to consider the problem from the perspective of Optimal Transport, rather than just maximization of the data log-likelihood, such that utilization of alternative latent-space constraints are justified. This will change the manner in which the aleatoric uncertainty is encapsulated by the latent densities.

\section{Methods}
\label{sec:methods}
It has been argued that the PU-Net can converge to sub-optimal latent representations due to ignoring information of the posterior latent distribution, which continuously inhibits gradient descent optimization. To approach this challenge, we present a general implementation that encourages latent space homogeneity. To address potentially inhomogeneous latent space variances, we propose the use of the Sinkhorn Divergence, which is strongly guided by entropic regularization and further encourages latent space homogeneity.

\subsection{Modifying the objective}
The ELBO can be rewritten as
\begin{subequations}
    \begin{align}
        \log \, p_{\bm{\phi}}(\mathbf{y}\mid\mathbf{x}) &\geq \mathbb{E}_{q_{\bm{\theta}}(\mathbf{z}\mid\mathbf{y},\mathbf{x})}\hspace{-3pt} \left[ \log \frac{p_{\bm{\phi}}(\mathbf{y},\mathbf{z}\mid\mathbf{x})}{q_{\bm{\theta}}(\mathbf{z}\mid \mathbf{y},\mathbf{x})} \right]  \notag \\
         % &= \mathbb{E}_{q_{\bm{\theta}}(\mathbf{z}\mid\mathbf{y},\mathbf{x})}\hspace{-3pt} \left[ \log p_{\bm{\phi}} (\mathbf{y}\mid \mathbf{x}, \mathbf{z}) + \log \frac{p_{\bm{\psi}}(\mathbf{z}\mid\mathbf{x})}{ q_{\bm{\theta}}(\mathbf{z}\mid\mathbf{x}) } + \log \frac{q_{\bm{\theta}}(\mathbf{y}\mid\mathbf{x})}{ q_{\bm{\theta}}(\mathbf{y}\mid\mathbf{z}, \mathbf{x})}\right] \notag \\
         &= \mathbb{E}_{q_{\bm{\theta}}(\mathbf{z}\mid\mathbf{y},\mathbf{x})}\hspace{-3pt} \left[ \log p_{\bm{\phi}} (\mathbf{y}\mid \mathbf{x}, \mathbf{z})\right. \notag \\ 
         &\hspace{10pt}\left.- \log \frac{q_{\bm{\theta}}(\mathbf{z}\mid\mathbf{x})}{ p_{\bm{\psi}}(\mathbf{z}\mid\mathbf{x}) } + \log \frac{q_{\bm{\theta}}(\mathbf{z}\mid\mathbf{x})}{ q_{\bm{\theta}}(\mathbf{z}\mid\mathbf{y}, \mathbf{x})}\right]  \tag{13}
    \end{align}
\end{subequations}
% Note how the point-wise mutual information is naturally minimized. Consider taking the expectation over the true dataset $p_\mathcal{D}(\mathbf{y}|\mathbf{x})$. We can write
% \begin{equation}
%     \log \, p_{\bm{\phi}}(\mathbf{y}\mid\mathbf{x})  \geq \mathbb{E}_{q_{\bm{\theta}}(\mathbf{z},\mathbf{y}\mid \mathbf{x})} \left[ \log p_{\bm{\phi}} (\mathbf{y}\mid \mathbf{x}, \mathbf{z})\right]  - \mathbb{E}_{p_{\mathcal{D}}(\mathbf{y}\mid \mathbf{x})}\operatorname{KL}[q_{\bm{\theta}}(\mathbf{z}\mid \mathbf{x}) || p_{\bm{\psi}} (\mathbf{z}\mid\mathbf{x})] - I(\mathbf{y},\mathbf{z}\mid\mathbf{x})
% \end{equation}
For the eventual segmentation task, good reconstruction and an \textit{aggregate} posterior, $q_{\bm{\theta}}(\mathbf{z}\mid \mathbf{x})$, congruent with the prior are essential. However, since both $q_{\bm{\theta}}(\mathbf{z}\mid \mathbf{x})$ and $p_{\bm{\psi}}(\mathbf{z}\mid \mathbf{x})$ are flexible, the KL divergence can be trivially minimized by collapsing both onto the same degenerate solution. As the KL divergence only measures density ratios, two distributions can reach zero divergence while occupying an arbitrarily small region of the latent space (Section~\ref{sec:challenges}). Hence, we rewrite the objective and introduce the hyperparameters $\alpha$ and $\beta$ similar to related literature~\cite{higgins2017beta}.  For a single annotation $\mathbf{y}$, we obtain
\begin{subequations}
    \begin{align}
        \mathcal{L}_{\mathbf{y}}&= -\mathbb{E}_{q_{\bm{\theta}}(\mathbf{z}\mid\mathbf{y},\mathbf{x})} \left[ \log p_{\bm{\phi}} (\mathbf{y}\mid \mathbf{x}, \mathbf{z})\right. \notag \\
        &\hspace{70pt} \left. - \beta\cdot\log \frac{ q_{\bm{\theta}}(\mathbf{z}\mid\mathbf{x}) }{p_{\bm{\psi}}(\mathbf{z}\mid\mathbf{x})} + \alpha\cdot\log\frac{q_{\bm{\theta}}(\mathbf{z}\mid\mathbf{x})}{q_{\bm{\theta}}(\mathbf{z}\mid\mathbf{y},\mathbf{x})}  \right]\notag \\
        &= -\mathbb{E}_{q_{\bm{\theta}}(\mathbf{z}\mid\mathbf{y},\mathbf{x})} \left[ \log p_{\bm{\phi}} (\mathbf{y}\mid \mathbf{x}, \mathbf{z}) \right. \notag \\
        &\hspace{70pt} \left. +\log \frac{q_{\bm{\theta}}(\mathbf{z}\mid\mathbf{x})^{\alpha-\beta}}{p_{\bm{\psi}}(\mathbf{z}\mid\mathbf{x})^{-\beta} q_{\bm{\theta}}(\mathbf{z}\mid\mathbf{y},\mathbf{x})^{\alpha}}  \right]  \notag \\
        &= -\mathbb{E}_{q_{\bm{\theta}}(\mathbf{z}\mid\mathbf{y},\mathbf{x})} \left[ \log p_{\bm{\phi}} (\mathbf{y}\mid \mathbf{x}, \mathbf{z}) \right. \notag \\
        &\hspace{70pt} \left. +\log \frac{q_{\bm{\theta}}(\mathbf{z}\mid\mathbf{x})^{\alpha-\beta}  q_{\bm{\theta}}(\mathbf{z}\mid\mathbf{y},\mathbf{x})^{-\alpha}}{p_{\bm{\psi}}(\mathbf{z}\mid\mathbf{x})^{\alpha - \beta} p_{\bm{\psi}}(\mathbf{z}\mid\mathbf{x})^{-\alpha}}  \right]  \notag
    \end{align}
\end{subequations}
Taking the expectation over annotations, where the aggregate posterior is $q_{\bm{\theta}}(\mathbf{z}\mid\mathbf{x})=\mathbb{E}_{p(\mathbf{y}\mid\mathbf{x})}[\,q_{\bm{\theta}}(\mathbf{z}\mid\mathbf{y},\mathbf{x})\,]$, we obtain the minimization objective
\begin{equation}
    \begin{aligned}
        \mathcal{L} &= -\mathbb{E}_{p(\mathbf{y}\mid\mathbf{x})}\mathbb{E}_{q_{\bm{\theta}}(\mathbf{z}\mid\mathbf{y},\mathbf{x})} \left[ \log p_{\bm{\phi}} (\mathbf{y}\mid \mathbf{x}, \mathbf{z}) \right] \\
        &\hspace{20pt}+ (\beta-\alpha) \cdot \operatorname{KL}[q_{\bm{\theta}}(\mathbf{z}\mid\mathbf{x})||p_{\bm{\psi}}(\mathbf{z}\mid\mathbf{x})] \\
        &\hspace{20pt}+ \alpha \cdot \mathbb{E}_{p(\mathbf{y}\mid\mathbf{x})}\operatorname{KL}[q_{\bm{\theta}}(\mathbf{z}\mid\mathbf{y},\mathbf{x})||p_{\bm{\psi}}(\mathbf{z}\mid\mathbf{x})].
    \end{aligned} \tag{14} \label{eq:maxmutinfelbo}
\end{equation}
In the next section, we will justify constraining the aggregate posterior $q_{\bm{\theta}}(\mathbf{z}\mid \mathbf{x})$ with the entropy-regularized Sinkhorn Divergence by taking the OT perspective on this segmentation problem.

\subsection{Sinkhorn Divergence on the aggregated posterior}
\label{subsec:spunet}
It has been shown in the previous section that the training objective can be modified to contain the KL-divergence with the aggregated posterior. By using the theoretical framework of Bousquet~\etal~\cite{bousquet2017optimal}, we will show that the KL constraint on the aggregated posterior can be an OT solution. Most works employ this technique to provide an alternative to the conventional VAE objective~\cite{patrini2019sinkhorn, tolstikhin2019wasserstein}. We specifically consider the optimization problem in the conditional setting for probabilistic segmentation and have framed previous literature according to this context.

Consider the random variables $(\mathbf{X},\*Y,\hat{\*Y},\*Z)\in\mathcal{X}\times\mathcal{Y}\times\mathcal{Y}\times\mathcal{Z}$, which correspond to the input image, ground-truth segmentation, model prediction and latent code, respectively. Then, we denote the joint distribution $P_{\*Y,\*Z\mid\*X}$, where a latent variable is obtained as $\*Z\sim P_{\*Z\mid \*X}$, and ground truth is obtained with $\*Y\sim P_{\*Y\mid \*Z,\*X}$. The OT problem considers couplings $\Gamma (\*Y,\hat{\*Y}\mid \*X)=\Gamma (\hat{\*Y} \mid \*Y, \*X) P(\*Y\mid\*X)$, where $\Gamma (\hat{\*Y}\mid \*Y, \*X)$ can be factored with a non-deterministic mapping through latent variable $\*Z$, as is shown by Bousquet~\etal~\cite{bousquet2017optimal}. In fact, considering a probabilistic encoder $Q_{\*Z\mid \*Y, \*X}$, deterministic decoder $f_{\bm{\phi}}$ and the set of joint marginals $\mathcal{P}_{\*Y, \*Z\mid \*X}$, the optimal solution of the OT problem can be stated as
\begin{subequations}
    \begin{align}
    W(&P_{\*Y\mid\*X},P_{\hat{\*Y}\mid\*X}) \notag \\
    &=\inf_{P\in\mathcal{P}_{\*Y,\*Z\mid\*X}} \mathbb{E}_{P_{\*Y,\*Z\mid\*X}} \left[c(\*Y,f_{\bm{\phi}}(\*Z,\*X)) \right]  \tag{15a}\label{eq:otsol1} \\
    &= \inf_{\mathcal{Q}:Q_{\*Z\mid\*X}=P_{\*Z\mid\*X}} \mathbb{E}_{P_{\*Y\mid\*X}}\mathbb{E}_{Q_{\*Z\mid\*Y,\*X}}\left[c(\*Y,f_{\bm{\phi}}(\*Z,\*X)) \right], \tag{15b} \label{eq:otsol2}
    \end{align}
\end{subequations}
with prior and aggregated posterior, $P_{\*Z\mid\*X}=P(\*Z\mid \*X)$ and $Q_{\*Z\mid\*X}=\mathbb{E}_{P_{\*Y\mid\*X}}\left[Q(\*Z\mid \*Y, \*X)\right]$, respectively (see Theorem 1 of Bousquet~\etal~\cite{bousquet2017optimal} for further details). Equation~(\ref{eq:otsol1}) follows by definition and Equation~(\ref{eq:otsol2}) indicates that rather than finding the couplings $\Gamma$, the search can be factored through probabilistic encoders $\mathcal{Q}$ which attempt to match the marginal $P_{\*Z\mid\*X}$. Subsequently, the constraint on the aggregated marginals can be relaxed with a convex penalty $D:\mathcal{Q}\times\mathcal{P}\rightarrow \mathbb{R}_+$ such that $D(Q_{\*Z\mid\*X},P_{\*Z\mid\*X})=0$ if, and only if $Q_{\*Z\mid\*X}=P_{\*Z\mid\*X}$. 

A viable option for this penalty is the Sinkhorn Divergence~\cite{cuturi2013sinkhorn}, which is convex, smooth and positive definite~\cite{feydy2019interpolating} and can serve as a surrogate for the KL-divergence on the aggregated posterior in Equation~(\ref{eq:maxmutinfelbo}). Furthermore, the entropic regularization can reduce the model's susceptibility to collapse to low-entropy solutions, as has been discussed in Section~\ref{subsec:OT}. The novel minimization objective, which is named the Sinkhorn PU-Net (SPU-Net) for reference, can be stated as
\begin{subequations}\label{eq:spunet-loss}
    \begin{align}
        % \mathcal{L} &= - \mathbb{E}_{q_\theta (\mathbf{z}\mid\mathbf{y},\mathbf{x})} [\,\log p_\phi (\mathbf{y}\mid\mathbf{z},\mathbf{x})\,] \,+ \notag \\
        %         & \hspace{25pt} (1-\alpha) \operatorname{KL} [\, q_\theta (\mathbf{z}\mid\mathbf{y},\mathbf{x}) || p_\psi (\mathbf{z}\mid\mathbf{x)} \,] +  \notag \\
        %         &\hspace{50pt} (\alpha + \beta - 1) \operatorname{S}_\epsilon [\,q_\theta (\mathbf{z}\mid\mathbf{x}) ||  p_\psi (\mathbf{z}\mid\mathbf{x} )\,],\tag{18}
        \mathcal{L} & = -\mathbb{E}_{p(\mathbf{y}\mid\mathbf{x})}\mathbb{E}_{q_{\theta}(\mathbf{z}\mid\mathbf{y},\mathbf{x})}[\log p_\phi (\mathbf{y}\mid\mathbf{x},\mathbf{z})] \notag \\
        &\hspace{40pt}+ \alpha \cdot \mathbb{E}_{p(\mathbf{y}\mid\mathbf{x})}\operatorname{KL}[q_{\bm{\theta}}(\mathbf{z}\mid \mathbf{y},\mathbf{x}) || p_{\bm{\psi}} (\mathbf{z}\mid\mathbf{x})]  \notag \\
        &\hspace{80pt}+ (\beta-\alpha) \cdot \operatorname{S}_\epsilon [q_{\bm{\theta}}(\mathbf{z}\mid \mathbf{x}) || p_{\bm{\psi}} (\mathbf{z}\mid\mathbf{x})], \tag{16}
    \end{align}
\end{subequations}
where $\alpha$ and $\beta$ are tunable parameters.
\begin{figure}
    \centering
    \includegraphics[width=0.45\textwidth]{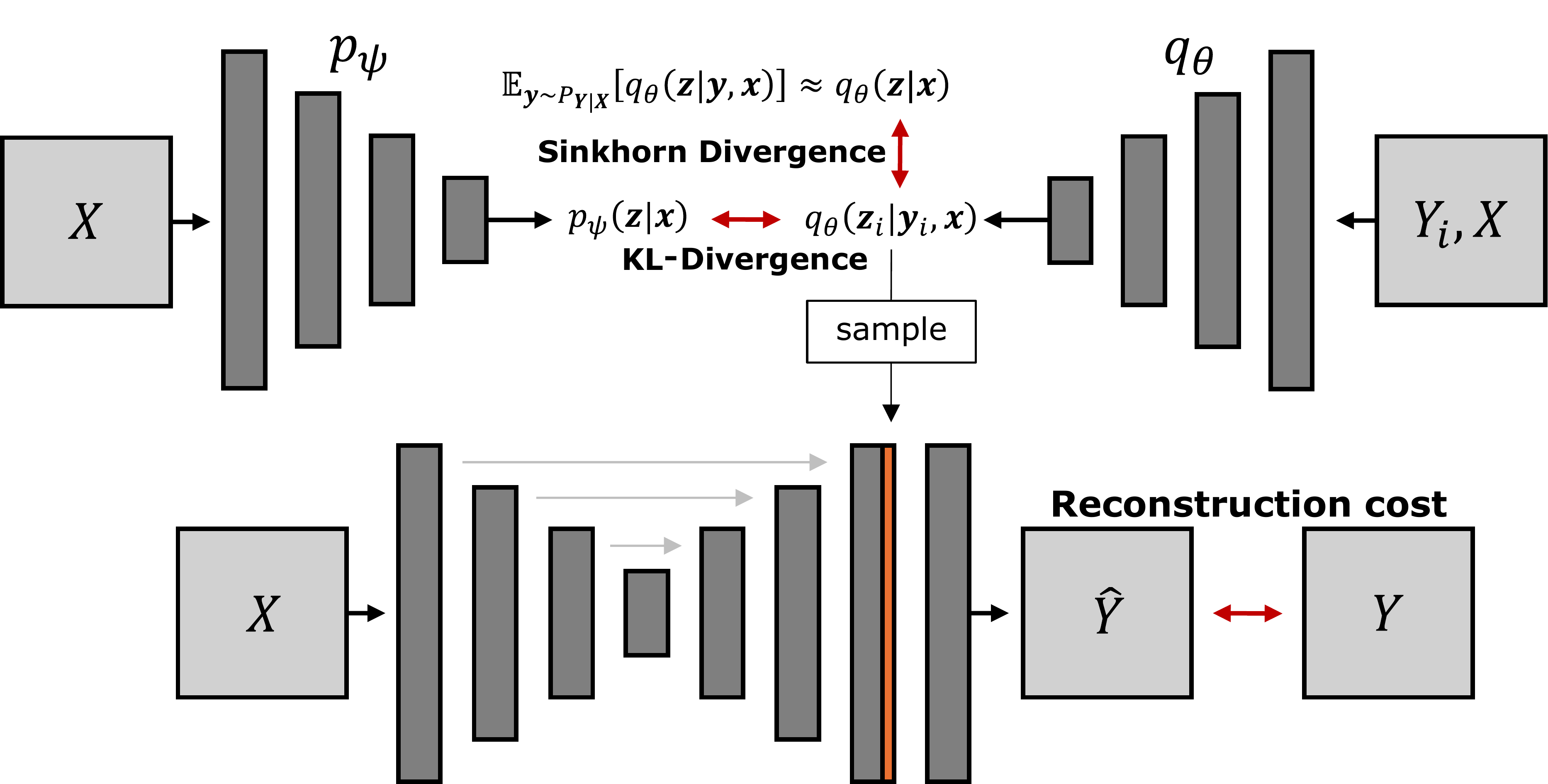}
    \caption{Schematic drawing of the Sinkhorn Probabilistic U-Net (SPU-Net) training framework, which is an extension of the PU-Net introduced by Kohl~\etal~\cite{kohl2018probabilistic}. The ground truth is denoted as $\mathbf{Y}$, the input image as $\mathbf{X}$, and the model prediction as $\mathbf{\hat{Y}}$. During testing, only the prior density $p_\psi (\mathbf{z}\mid \mathbf{x})$ is used to predict samples.}\label{fig:punets}
\end{figure}
\subsection{Data}
This study employs a diverse range of post-processed, multi-annotated medical image segmentation datasets, which exhibit significant variations in the ground truth, to train and evaluate the proposed models. Furthermore, the utilization of publicly available datasets underscores the clinical significance of this research. Firstly, we use the LIDC-IDRI~\cite{lidc} dataset, a popular benchmark for aleatoric uncertainty quantification, which contains lung-nodule CT scans. Secondly, we use the Pancreas, Pancreatic-lesion CT datasets provided by the QUBIQ~2021~Challenge~\cite{qubiq2021}. See Table~\ref{tab:datadetails} for an overview. Some datasets of the QUBIQ challenge are omitted due to their tiny sizes, which make training unstable and evaluation extremely sensitive to a few test samples, inhibiting meaningful analysis.

In line with related literature, it is assumed that the region of interest is detected and the desired uncertainty is solely regarding its exact delineation. This is also common practise in clinical procedures, where lesions are first detected and only then characterized more precisely~\cite{schreuder2021artificial,ciompi2017towards,lam2021contemporary,lemense2020development,jazieh2018saudi}. Therefore, each image is center-cropped to 128$\times$128-pixel patches. Furthermore, a train/validation/test set is exploited where approximately 20\% of the dataset is reserved for testing and three-fold cross-validation is used on the remaining 80\%. We carefully split the data according to patient number to avoid data leakage or model bias.
\begin{table}
\centering
\setlength{\tabcolsep}{4pt}
{\resizebox{\linewidth}{!}{
\begin{tabular}{c|l|ccccc} 
\toprule
\multicolumn{2}{c}{\textbf{Dataset}} & \textbf{Modality} & \textbf{Annotators} & \textbf{Samples} &   \\
\midrule
\multicolumn{2}{c|}{LIDC-IDRI} & CT& $4$ & $15096$   \\
\midrule 
\multirow{2}{*}{QUBIQ} & Pancreas           &CT     & $2$ & $702$   \\
                        & Pancreatic lesion &CT     & $2$ & $206$ \\
\bottomrule 
\end{tabular}}
    \caption{Modality, number of annotators and size for the used datasets.}\label{tab:datadetails}}
\end{table}

\subsection{Training}
\label{subsec:training}
The proposed SPU-Net is compared to the PyTorch implementation of the PU-Net~\cite{kohl2018probabilistic} and its NF-augmented variant (with a 2-step planar flow)~\cite{valiuddin2021improving}, which we will refer to as the PU-Net+NF. We point readers to the respective papers for further details on both the PU-Net and PU-Net+NF architectures, which also serve as the foundation of the SPU-Net.

Optimizing Equation~(\ref{eq:spunet-loss}) requires the aggregate posterior, i.e. integration over all segmentation masks. The aggregated posterior can be approximated as $q_\theta (\mathbf{z}\mid\mathbf{x})\approx\mathbb{E}_{p_{\mathbf{Y}\mid\mathbf{X}}} [\,q_\theta (\mathbf{z}\mid \mathbf{y}, \mathbf{x})\,]$. A sample of each posterior is taken and their Sinkhorn Divergence with the equivalent number of samples from the prior is determined. A randomly selected posterior is used for the reconstruction loss (cross-entropy) and analytical KL divergence with the input conditional prior $p_{\bm{\psi}}(\mathbf{z}\mid\mathbf{x})$. The training procedure is shown with pseudocode in Algorithm~\ref{algo:psuedo}.
% Our implementation of the PU-Net consists of four-layer deep prior and posterior networks, where each layer consists of three sequential convolutions with 32, 64, 128, 256 filters, respectively. The encoding-decoding U-Net is also four layers deep and has identical filters. Latent samples from the encoding networks are up-scaled with a channel-wise tiling procedure to match the feature map dimenionsality of the U-Net. Both the U-Net feature map and the up-scaled latent vector are then concatenated, and are subjected to four $1\times1$-convolutions before being activated by a Sigmoid function. The network weights are initialized according to the Kaiming procedure~\cite{he2015delving} and the biases are sampled from a truncated normal with zero mean and a standard deviation of $0.001$. 

For each dataset, we train the PU-Net, PU-Net+NF and our proposed method. Each of those models is trained on three folds using cross-validation. Furthermore, the experiments per model and dataset are conducted across four latent dimensionalities. All hyperparameters across experiments are kept identical, except for the number of training epochs and parameters related to the Sinkhorn Divergence. This implies across all datasets a batch size of $32$, using the Adam optimizer with a maximum learning rate of $10^{-4}$ that is scheduled by a linear warm-up of 5 and 50 epochs for the LIDC-IDRI and QUBIQ datasets, respectively, cosine annealing, and weight decay of $10^{-5}$. For our proposed method, we additionally constrain the gradient norm to one. For the LIDC-IDRI dataset we set the number of training epochs to 200 and for all QUBIQ datasets we trained for 2,500 epochs. After training, we used the model checkpoint with the lowest validation loss. This was usually much earlier than the set number of epochs. To implement the Sinkhorn Divergence, we make use of the \texttt{GeomLoss}~\cite{feydy2019interpolating} package. 

Furthermore, data augmentation has been used during training, which consisted of random rotations within $[-180, 180]$ degrees, scaling within $[0.8, 1.2]$, and translation and shears in all directions of $0.1$ and $30\%$, respectively. The number of samples taken from the densities equals the number of available segmentation masks from each dataset.
\begin{algorithm}
\caption{Pseudocode of the proposed training algorithm}
\label{algo:psuedo}
$\mathcal{Y} \sim p_{\mathbf{Y}\mid\mathbf{X}}$\;
$k \gets \text{random.int}(n\_masks)$\;
\ForAll{$\mathbf{y}_i \in \mathcal{Y}$}{
    $\mathbf{z}_i^\theta \sim q_\theta(\mathbf{z} \mid \mathbf{y}_i,\mathbf{x})$\;
    $\mathbf{z}_i^\psi \sim p_\psi(\mathbf{z} \mid \mathbf{x})$\;
    }
$\mathbf{\hat{y}} \sim p_\phi (\mathbf{y} \mid \mathbf{x}, \mathbf{z}_k^\theta)$\;
$loss \gets \text{CE}(\mathbf{y}_k, \mathbf{\hat{y}}) + \alpha\cdot\text{KL}(q_\theta(\mathbf{z}\mid\mathbf{y}_k,\mathbf{x}) \,\|\, p_\psi(\mathbf{z}\mid\mathbf{x})) + (\beta-\alpha)\cdot\text{Sinkhorn}(\mathbf{z}^\theta, \mathbf{z}^\psi)$\;
\end{algorithm}
\subsection{Evaluation}
\label{subsec:evaluation}
\subsubsection{Data distribution}
To evaluate the model performances, Hungarian Matching has recently been used as an alternative to the Generalized Energy Distance (GED). This is because the GED can be biased to simply reward sample diversity when subject to samples of poor quality~\cite{kohl2019hierarchical,valiuddin2021improving}. We refer to Hungarian Matching as the Empirical Wasserstein Distance, abbreviated as EWD or $\hat{W}_{k}$, since it equals the Wasserstein distance between the discrete sets of samples from the model and the ground-truth distribution. 

We implement the EWD as follows. Each set of ground-truth segmentations is replicated to match the number of predictions $N$ obtained during testing. We then apply a cost metric $k$ to each combination of elements in the ground-truth and prediction set, to obtain an $N\times N$ cost matrix. Subsequently, the unique optimal coupling between the two sets that minimizes the average cost is determined. We use unity minus the Intersection over Union (1 - IoU) as cost function for $k$. Correct empty predictions are assigned zero cost. We sample 16 predictions for the LIDC-IDRI, Pancreas and Pancreatic-lesion datasets, which were specifically chosen to be multiples of the number of available annotations.

\subsubsection{Latent distribution}
The method used to determine the latent space homogeneity is the Effective Rank~(ER)~\cite{roy2007effective}, which can be regarded as an extension of the conventional matrix rank. The ER is defined as
\begin{equation}
    \operatorname{ER}(\bm{\sigma})=e^{-\sum c_i \log c_i},
\end{equation}
where $c_i$ are elements of the normalized singular values $\mathbf{c}=\frac{\bm{\sigma}}{||\bm{\sigma}||_1}$. For this approach, the singular value vector $\bm{\sigma}=[\sigma_1, \sigma_2,...,\sigma_d]$ needs to be determined. Since the densities are axis-aligned Normals, $\bm{\sigma}$ follows directly from the diagonal of the covariance matrix, and no Singular Value Decomposition is required.

\section{Results \& Discussion}
\label{sec:results}
\begin{table*}
\centering
\setlength{\tabcolsep}{12pt}
{\resizebox{\linewidth}{!}{
\begin{tabular}{@{\hspace{5pt}}c@{\hspace{5pt}}|@{\hspace{5pt}}c@{\hspace{5pt}}|@{\hspace{5pt}}c@{\hspace{5pt}}c@{\hspace{10pt}}c@{\hspace{5pt}}c@{\hspace{10pt}}c@{\hspace{5pt}}c@{\hspace{10pt}}c@{\hspace{5pt}}c@{\hspace{5pt}}c@{\hspace{5pt}}} 
\toprule
\multirow{2}{*}{\underline{Dataset}}  & \multirow{2}{*}{\underline{Model}}  &   \multicolumn{2}{c}{$\mathbf{d=4}$} &  \multicolumn{2}{c}{$\mathbf{d=5}$}    & \multicolumn{2}{c}{$\mathbf{d=6}$} & \multicolumn{2}{c}{$\mathbf{d=7}$} \\ 
& & $\hat{W}_{k}\downarrow$  & ER$\uparrow$ & $\hat{W}_{k}\downarrow$ &  ER$\uparrow$ & $\hat{W}_{k}\downarrow$  & ER$\uparrow$ & $\hat{W}_{k}\downarrow$  &  ER$\uparrow$ \\
\midrule
             
                \multirow{3}{*}{LIDC-IDRI}         
                &PU-Net     & $0.451\pm0.001$           & $1.68\pm0.07$           & $0.450\pm0.003$           &  $1.68\pm0.08$ & $0.451\pm0.006$                & $1.92\pm0.05$           & $0.451\pm0.002$  &  $1.77\pm0.14$ \\
                &PU-Net+NF  & $0.451\pm0.004$           & $2.90\pm0.39$           & $0.450\pm0.002$           &  $3.31\pm0.67$  & $0.446\pm0.003$      & $2.94\pm0.59$  & $0.450\pm0.002$           &  $4.00\pm0.52$         \\
                &Ours    & $0.447\pm0.005$           & $3.52\pm0.04$           & $0.441\pm0.002$           &  $4.57\pm0.03$  & $\mathbf{0.440\pm0.006}$     & $5.12\pm0.10$  & $0.443\pm0.002$           &  $5.59\pm0.06$       \\                     
\midrule 
                \multirow{3}{*}{\makecell{QUBIQ \\ Pancreas}} 
                &PU-Net     & $0.602\pm0.034$           & $1.76\pm0.07$           & $0.608\pm0.024$ &  $1.49\pm0.25$    & $0.603\pm0.030$    & $1.52\pm0.23$  & $0.589\pm0.002$  &  $1.65\pm0.22$          \\
                &PU-Net+NF  & $0.581\pm0.004$  & $2.37\pm0.24$  & $0.610\pm0.080$ &  $3.15\pm0.43$    & $0.585\pm0.021$    & $2.77\pm0.24$  & $0.589\pm0.054$           &  $3.77\pm0.98$         \\
                &Ours    & $0.541\pm0.011$ &  $3.44\pm0.07$  & $0.532\pm0.004$    & $4.13\pm0.24$  & $\mathbf{0.530\pm0.016}$    & $5.33\pm0.08$  & $0.536\pm0.011$           &  $6.52\pm0.18$        \\  
\midrule
                \multirow{3}{*}{\makecell{QUBIQ \\ Pancreatic \\ lesion}}  
                &PU-Net     & $0.628\pm0.009$  & $1.25\pm0.11$ & $0.606\pm0.004$    &  $1.18\pm0.03$  & $0.618\pm0.002$            & $1.07\pm0.02$           & $0.611\pm0.007$           &  $1.31\pm0.16$          \\
                &PU-Net+NF  & $0.536\pm0.024$  & $2.54\pm0.35$ & $0.559\pm0.022$             &  $3.21\pm0.30$           & $0.568\pm0.063$            & $3.06\pm0.29$           & $0.535\pm0.025$  &  $4.88\pm0.43$  \\
                &Ours    & $0.537\pm0.018$  & $3.52\pm0.11$ & $0.539\pm0.000$             &  $4.41\pm0.28$           & $\mathbf{0.518\pm0.039}$   & $4.20\pm0.34$  & $0.543\pm0.011$            &  $5.33\pm0.23$        \\  

\bottomrule
\end{tabular}}
    \caption{Comparison of the experimented models on various datasets. The PU-Net is re-implemented from~\cite{kohl2018probabilistic} and PU-Net+NF from~\cite{valiuddin2021improving}. Results are evaluations on the test set. The mean and standard deviations (threefold cross-validation) of the Empirical Wasserstein metric with cost unity minus the Intersection over Union (1-IoU) are presented. Bold indicates the best $\hat{W}_{k}$ per dataset. Furthermore, the homogeneity of the latent singular values is expressed with the Effective Rank. A clear relationship between a homogeneous latent space and model performance is visible.}\label{tab:results}}
\end{table*}

We confirm our implementation of the baseline PU-Net with several experiments. When evaluating the PU-Net using the LIDC-IDRI dataset, we have obtained GED values of 0.327$\,\pm\,$0.003, which closely align with the reported values in the work of Kohl~\etal~\cite{kohl2019hierarchical}, yet exhibit significantly reduced standard deviation. Moreover, the $\hat{W}_{k}$ values surpass the values reported by Kohl~\etal~\cite{kohl2019hierarchical}. These discrepancies can be potentially attributed to differences in training splits and model initialization. Nonetheless, these findings underscore the comparable performance of our experiments, warranting a reliable basis for accurate comparison.

All models were trained on a 24GB NVIDIA RTX 3090. We show relevant statistics related to model complexity in Table~\ref{tab:stats}. Our methodology leaves inference time, memory usage and parameter count nearly unchanged, but increases the training time by approximately $1.75\times$ and the number of floating-point operations (FLOPs) by approximately $2.9\times$ w.r.t. the baseline PU-Net. This is expected since our algorithm evaluates multiple posteriors in a training step. While training complexity is not the focus of this work, it is likely that evaluation in this manner can potentially reduce the number of epochs required to fully optimize the model. This is because random sampling is an approximation of the aggregated posterior.
\begin{table}[!h]
\centering
{\resizebox{\linewidth}{!}{
\begin{tabular}{l|ccccc} 
\toprule
\multirow{2}{*}{\textbf{Model}} & \textbf{Training time} & \textbf{Inference time } & \textbf{Memory} & \textbf{Parameters} & \textbf{FLOPs} \\
  & (appr. min/epoch) & (100$\times$, s.) &  (GB) & (M) & (G) \\
\midrule
PU-Net & 1.00     & 2.56 &  3.98  & 6.844 & 119 \\
PU-Net+NF & 1.13  & 2.57 &  4.07  & 6.853 & 211  \\
Ours        &  1.75 & 2.57 & 4.07 & 6.846 & 350 \\
\bottomrule
\end{tabular}}}
    \caption{Training statistics for the used models with batch size of 32. While training costs are increased, note that our methodology adds negligible costs to inference time, memory usage and parameter count.}\label{tab:stats}
\end{table}

The quantitative results from the conducted experiments are presented in Table~\ref{tab:results}. Values within the range of 4\,$\leq$\,$d$\,$\leq$\,7 have been used for the experiments. We have found values $\alpha\,$=$\,10$ and $\beta\,$=$\,110$ to sufficiently reduce the latent sparsity during training. For each dataset, the best performing model across the experimented latent dimensionalities is indicated in boldface. It can immediately be observed that the proposed method performs best in 10 of the 12 settings and achieves the best overall score on every dataset in terms of aleatoric uncertainty quantification (visible from the EWD metric). Additionally, a positive correlation is apparent between the latent space homogeneity (quantified with the ER) and performance. The algorithm evenly distributes the variances across all latent dimensionalities, which results in the best performance in terms of the EWD. This is in contrast to the PU-Net, which has a sparse latent space (i.e. low ER) and generally performed worse. The results for the PU-Net+NF support this empirical correlation as well, since its metric evaluations often reside between the former two in terms of both performance and latent space sparsity.

It is noteworthy that our methodology improves over previous literature with smaller latent dimensionalities, as this constrains its capacity. For example, our method applied on the Pancreas dataset with $d\,$=$\,4$ surpasses the PU-Net and PU-Net+NF with values $d\,$$\geq$$\,4$. For the other two datasets, our method with $d\,$=$\,4$ outperforms the PU-Net at every dimensionality, while the PU-Net+NF reaches comparable values at some dimensionalities. This serves as evidence that the information is better captured in the latent variances of the proposed method. Furthermore, we see that the performance slightly degrades above optimal $d$, which has also been discussed in Dai~\etal~\cite{dai2019diagnosing}. Nonetheless, our model outperforms the former approaches at nearly every dimensionality.

\begin{figure}[!tbp]
\centering
\begin{subfigure}{.42\textwidth}
\centering
  \includegraphics[width=\linewidth]{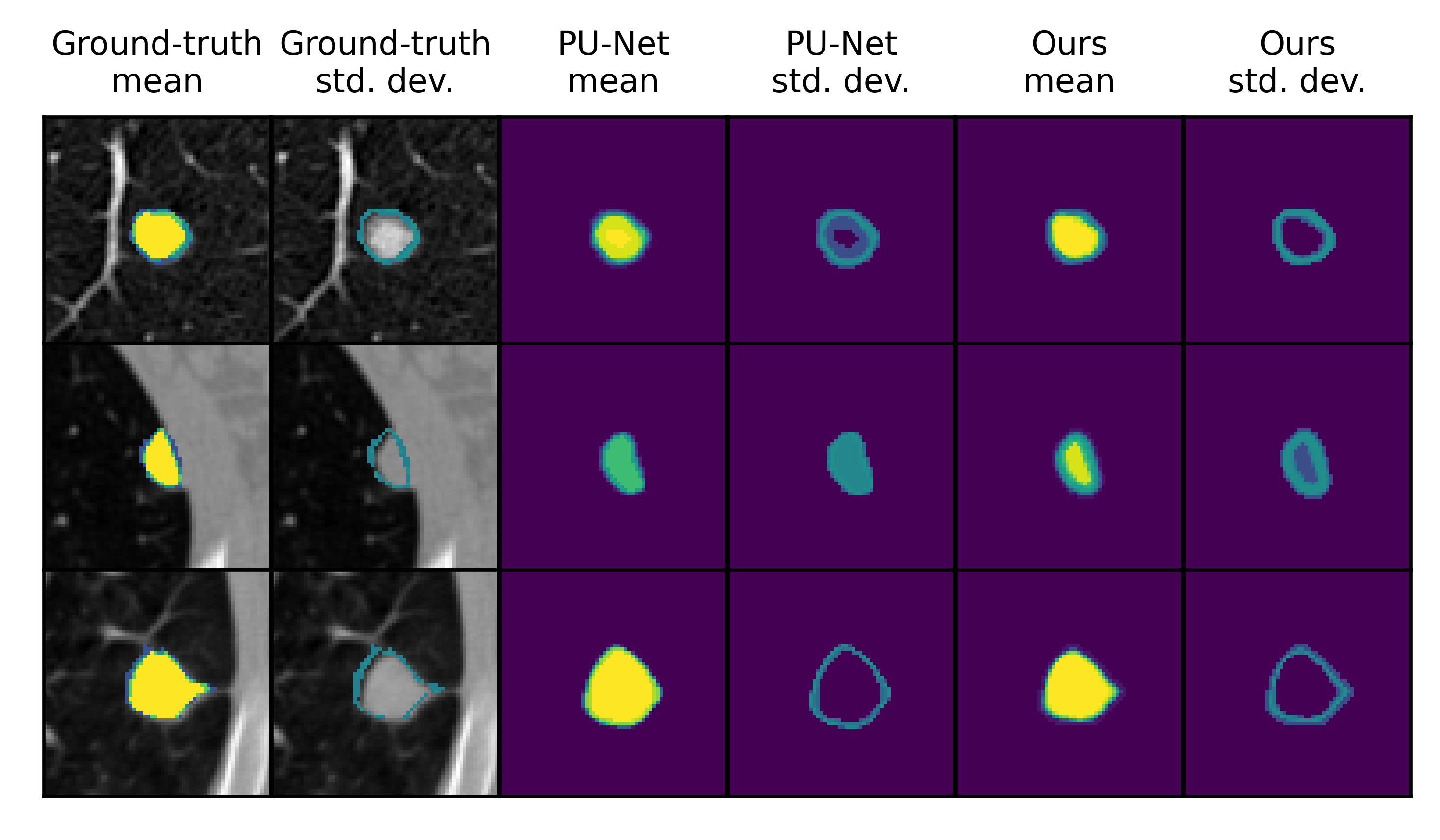}
  \caption{LIDC-IDRI}
  \label{fig:sub1}
\end{subfigure}%
\hspace{10pt}
\begin{subfigure}{.42\textwidth}
\centering
  \includegraphics[width=\linewidth]{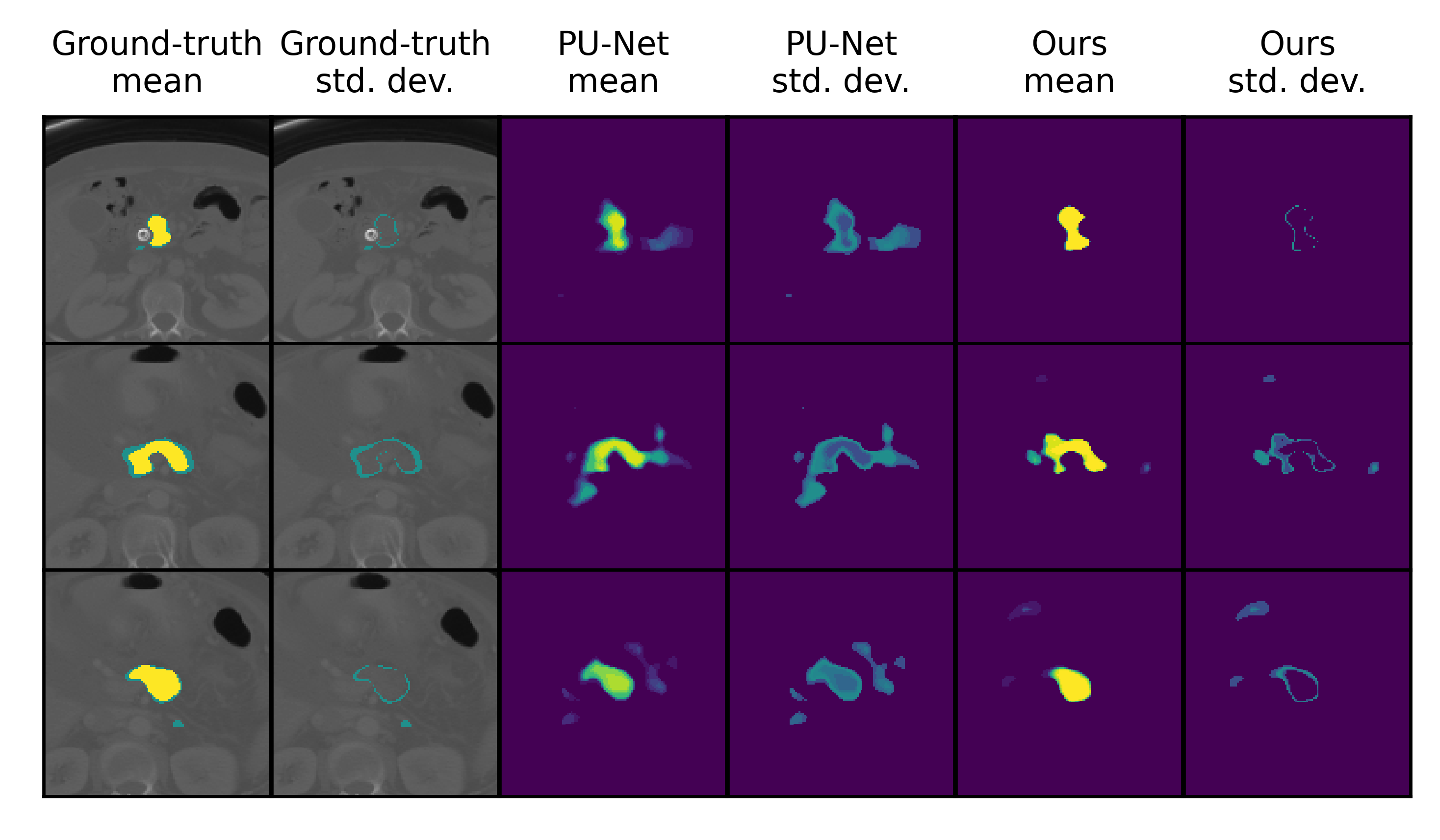}
  \caption{QUBIQ Pancreas}
  \label{fig:sub2}
\end{subfigure}
\hspace{10pt}
\begin{subfigure}{.42\textwidth}
\centering
  \includegraphics[width=\linewidth]{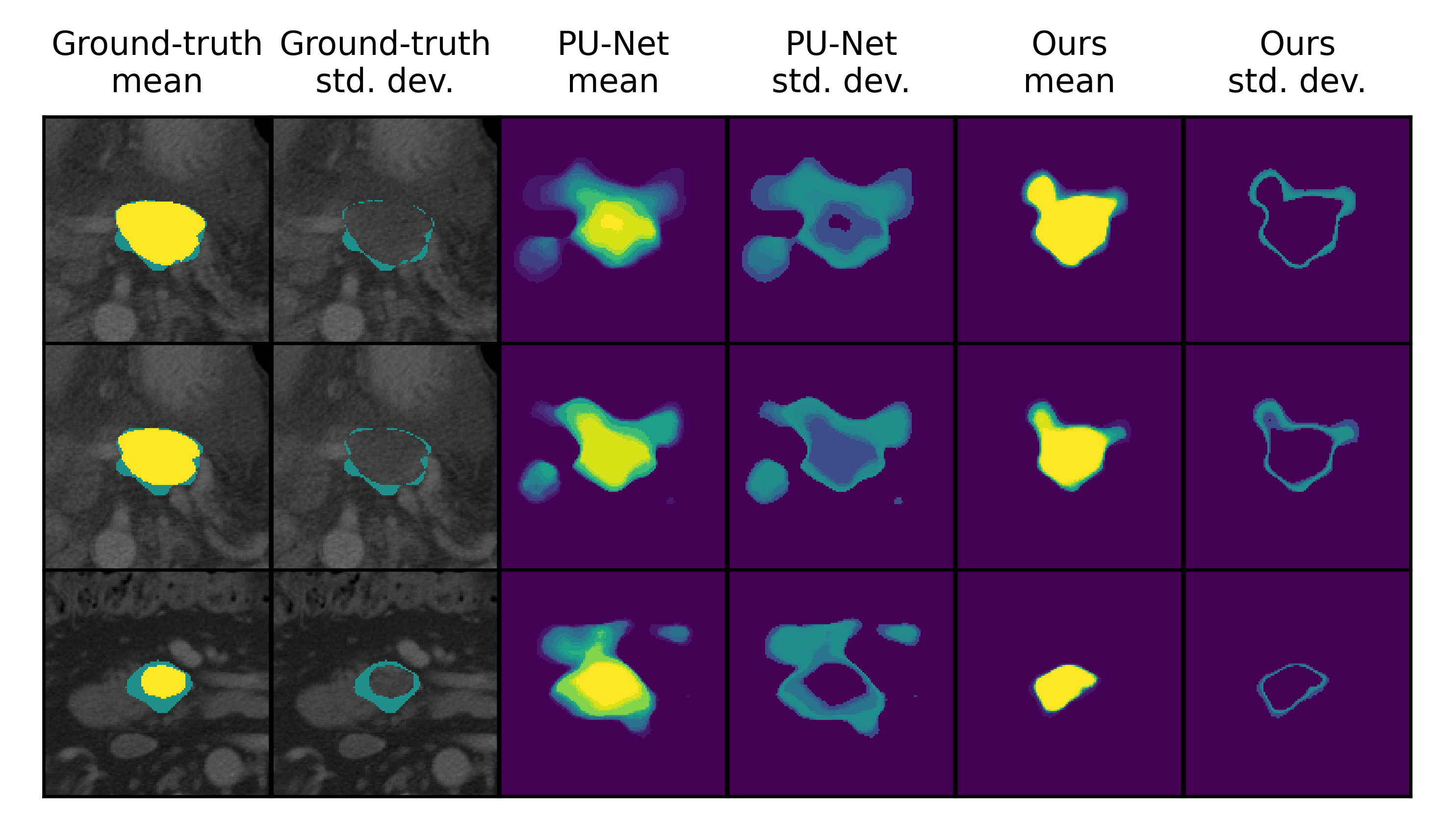}
  \caption{QUBIQ Pancreatic lesion}
  \label{fig:sub2}
\end{subfigure}
\caption{Visualization of test set predictions. The mean and standard deviation is taken from 16 predictions for both PU-Net and our proposed model.}
\label{fig:samplesall}
\end{figure}

While the quantitative evaluation confirms the superior performance, clinical setting deployment will mainly entail visual evaluation. Therefore, we qualitatively inspect the mean and standard deviation of the sampled predictions of the (S)PU-Net for the various datasets in Figure~\ref{fig:samplesall}. From the figure, the improvements in predictions over the baseline models are clearly visible. The predicted uncertainty closely matches the ground-truth, as discernible around borders of the prediction ensemble in the LIDC-IDRI dataset. For the Pancreas and Pancreatic-lesion datasets, large improvements in the mean predictions compared to the baselines can be observed. As expected, the lower EWD scores are clearly reflected in the sampled segmentation masks. These results expose the unfavourable latent space of the PU-Net.

\begin{figure}[!tbp]
\centering
\begin{subfigure}{.43\textwidth}
\centering
  \includegraphics[width=\linewidth]{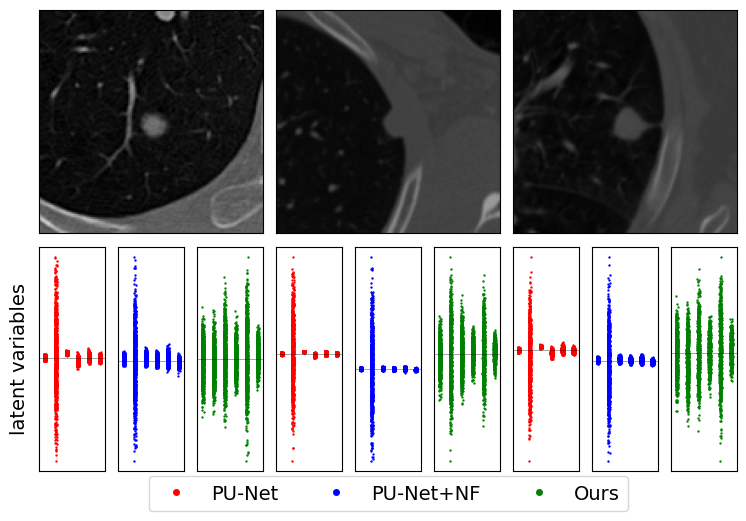}
  \caption{LIDC-IDRI}
  \label{fig:sub1}
\end{subfigure}%
\hspace{10pt}
\begin{subfigure}{.43\textwidth}
\centering
  \includegraphics[width=\linewidth]{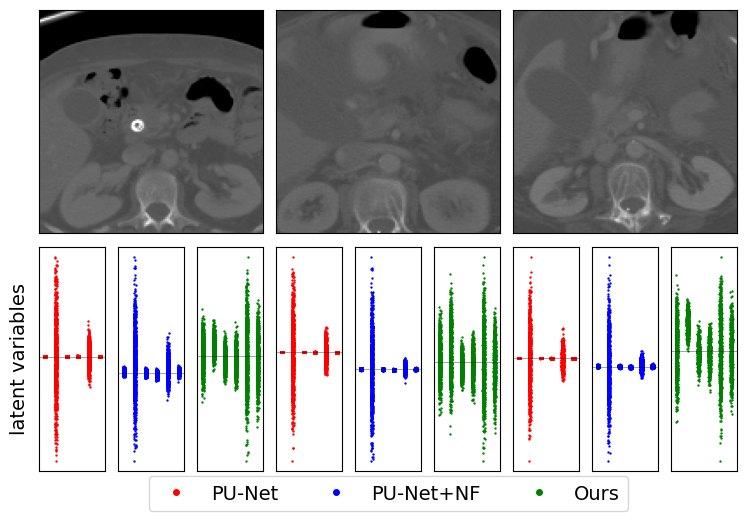}
  \caption{QUBIQ Pancreas}
  \label{fig:sub2}
\end{subfigure}
\hspace{10pt}
\begin{subfigure}{.43\textwidth}
\centering
  \includegraphics[width=\linewidth]{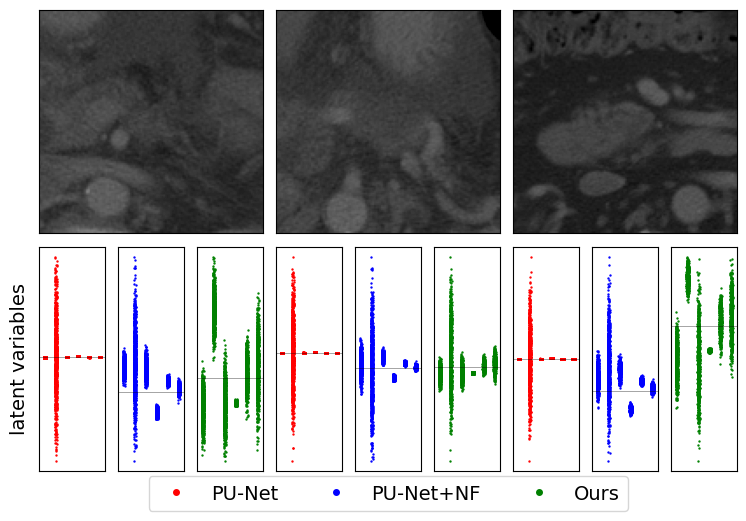}
  \caption{QUBIQ Pancreatic lesion}
  \label{fig:sub2}
\end{subfigure}
\caption{Visualization of the latent vector mean and variances subject to various test images.}
\label{fig:allvars}
\end{figure}

To better understand the distribution of the latent dimensions, the mean and variances of the image-conditional prior densities are depicted in Figure~\ref{fig:allvars}. Here, it can clearly be seen that the PU-Net latent representation is severely sparse. In line with the previously discussed results, it can be observed that the SPU-Net has the most homogeneous prior latent density and that the mean values are more purposeful since they are not all centered around zero.

\begin{figure}[!tbp]
\centering
\begin{subfigure}{.4\textwidth}
\centering
  \includegraphics[width=\linewidth]{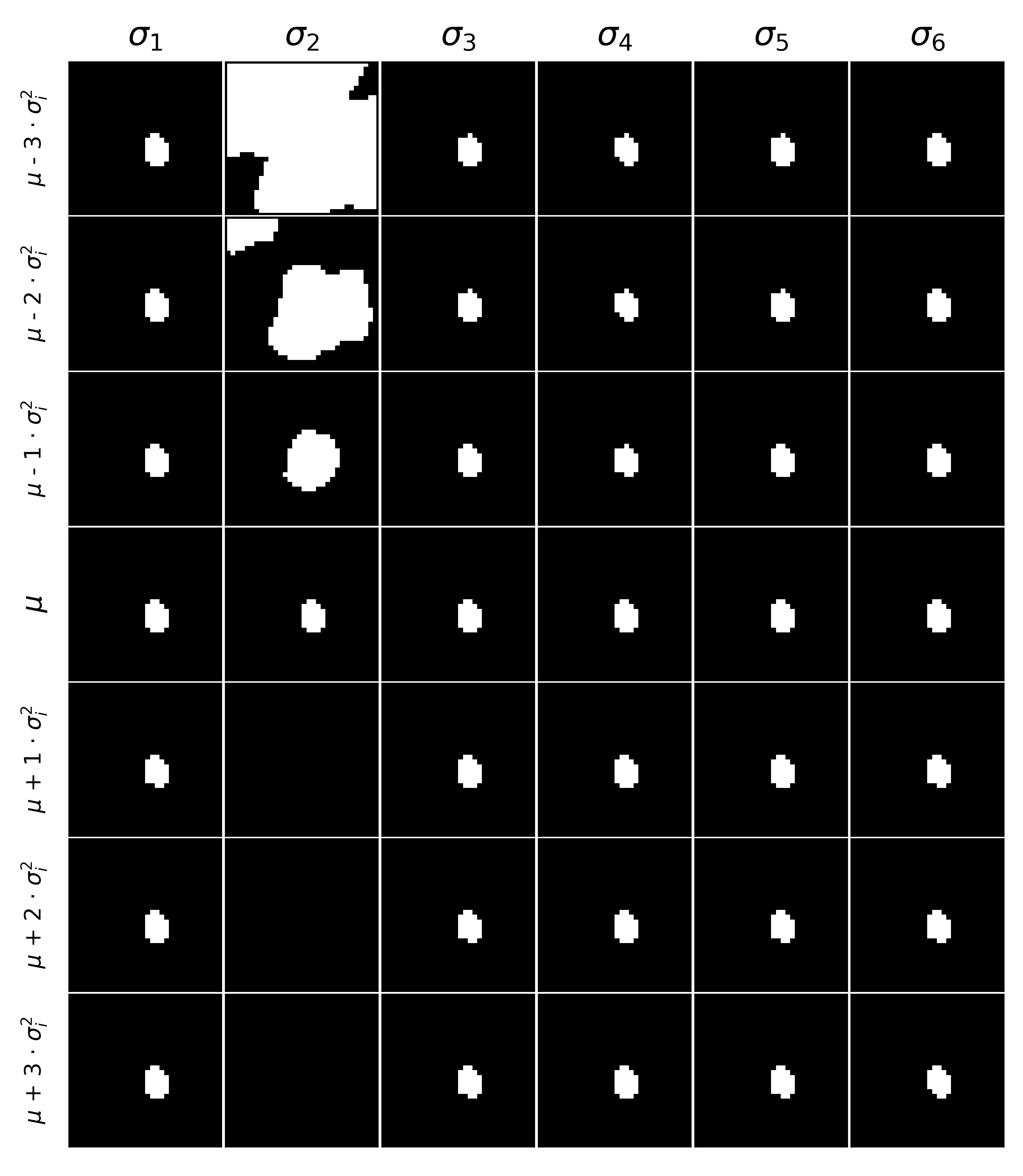}
  \caption{PU-Net}
  \label{fig:sub1}
\end{subfigure}%
\hspace{10pt}
\begin{subfigure}{.4\textwidth}
\centering
  \includegraphics[width=\linewidth]{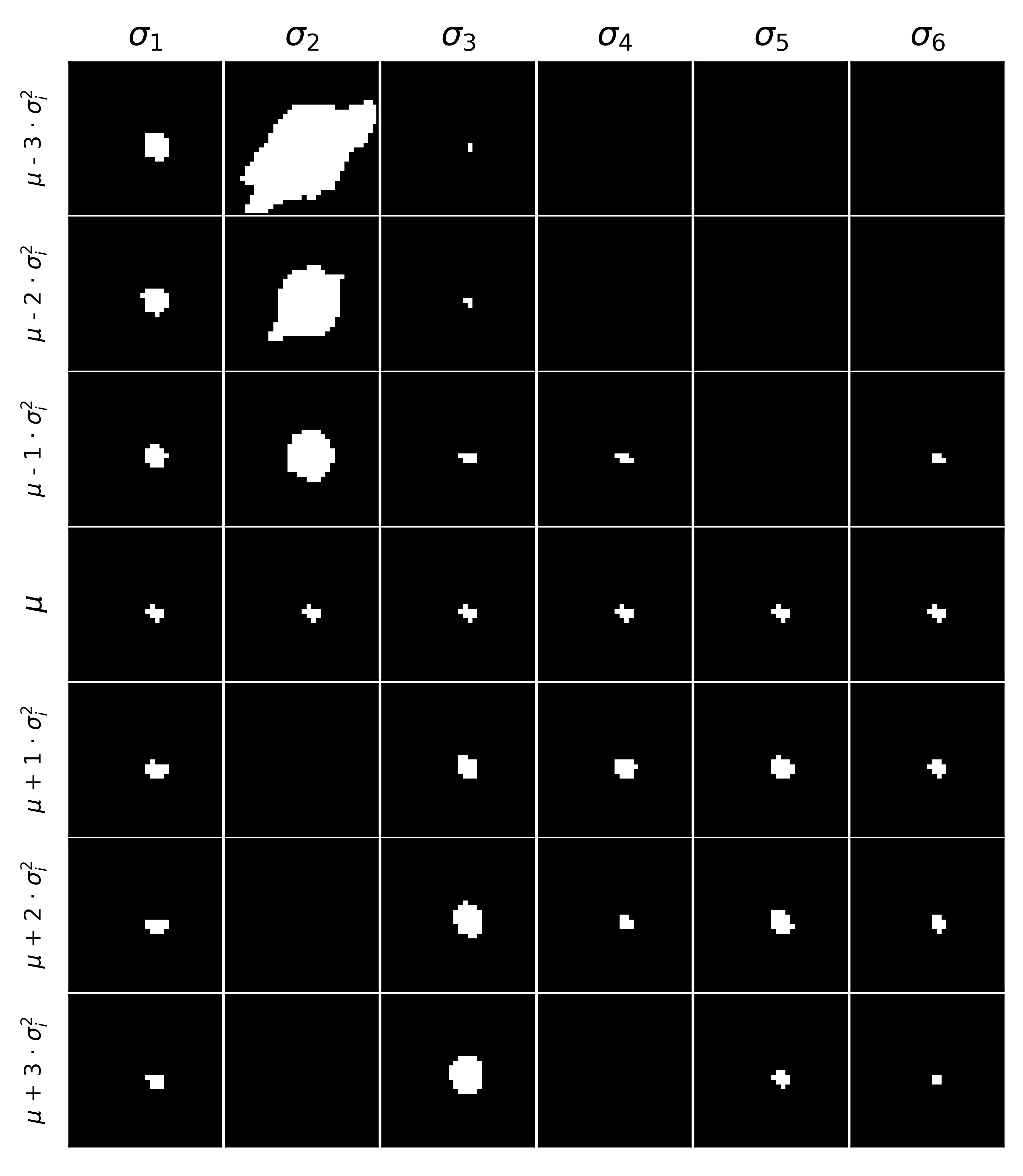}
  \caption{Ours}
  \label{fig:sub2}
\end{subfigure}
\caption{Interpolation of the latent axis-aligned prior density for the LIDC-IDRI dataset (zoomed-in for viewing convenience).}
\label{fig:collapsetotalabl}
\end{figure}

The effect of our modified training objective is confirmed in Figure~\ref{fig:collapsetotalabl}. Interpolation across latent dimensions almost has no effect on the PU-Net. Interestingly enough, the figure reflects at most two active latent dimensions, which reflects the obtained ER of $1.92$. In the proposed method, it can be observed that at least five latent dimensions influence the output, which is also in line with the found result ER$=\,$$5.12$. This indicates that the latent space has much more influence on the output of the model.

To further investigate latent behaviour, we sequentially `freeze' latent dimensions by taking their mean rather than sampling from the density in Figure~\ref{fig:ablationvars}. If the variance contributes to the output, then the model performance should degrade as dimensions are frozen. This clearly holds for the proposed method, but not for the PU-Net. The diagram also reflects the found ER values in Table~\ref{tab:results}.

\begin{figure}
\centering
\begin{subfigure}{.45\textwidth}
\centering
  \includegraphics[width=\linewidth]{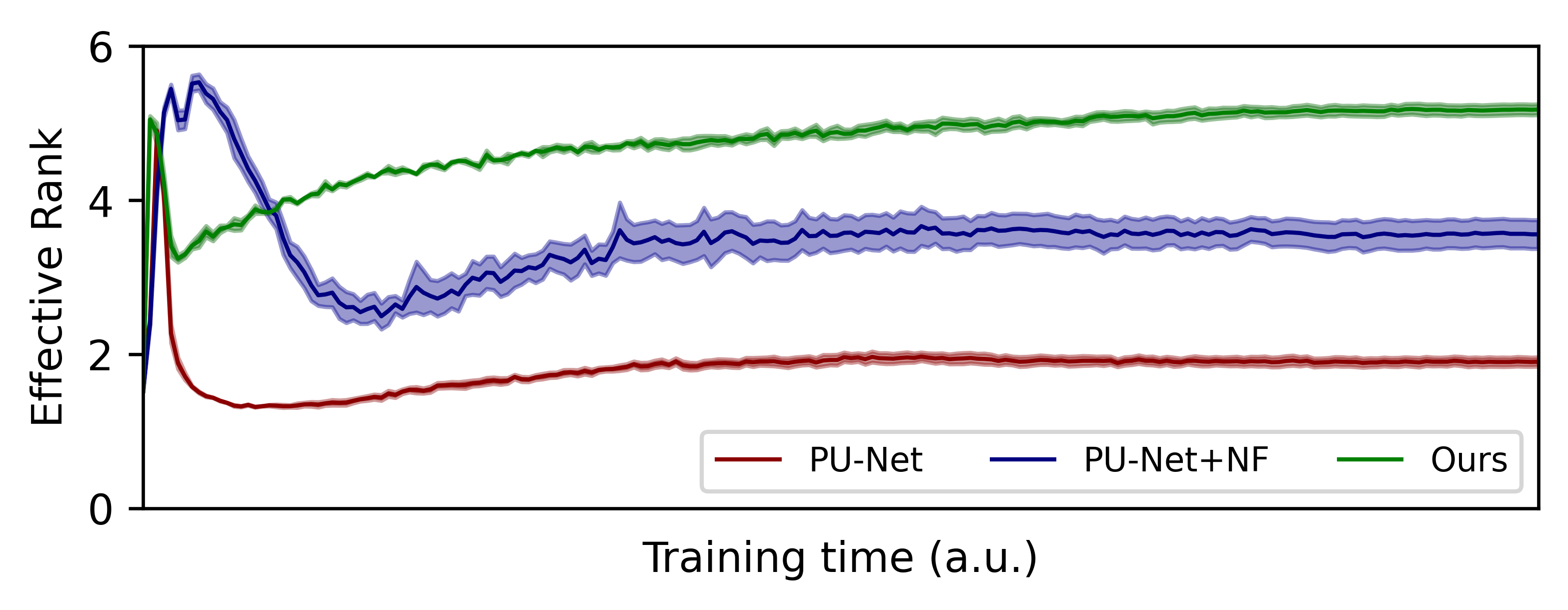}
  \caption{LIDC IDRI}
  \label{fig:sub1}
\end{subfigure}%
\hspace{10pt}
\begin{subfigure}{.45\textwidth}
\centering
  \includegraphics[width=\linewidth]{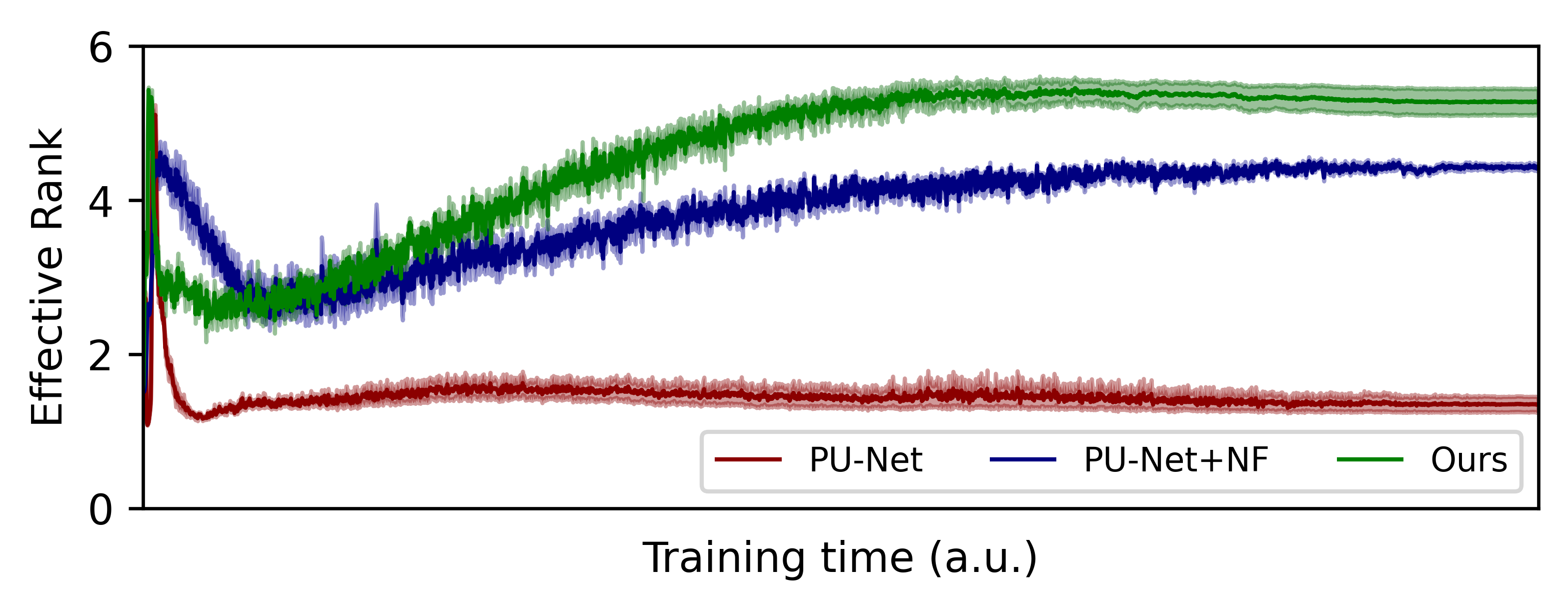}
  \caption{QUBIQ pancreas}
  \label{fig:sub2}
\end{subfigure}
\hspace{10pt}
\begin{subfigure}{.45\textwidth}
\centering
  \includegraphics[width=\linewidth]{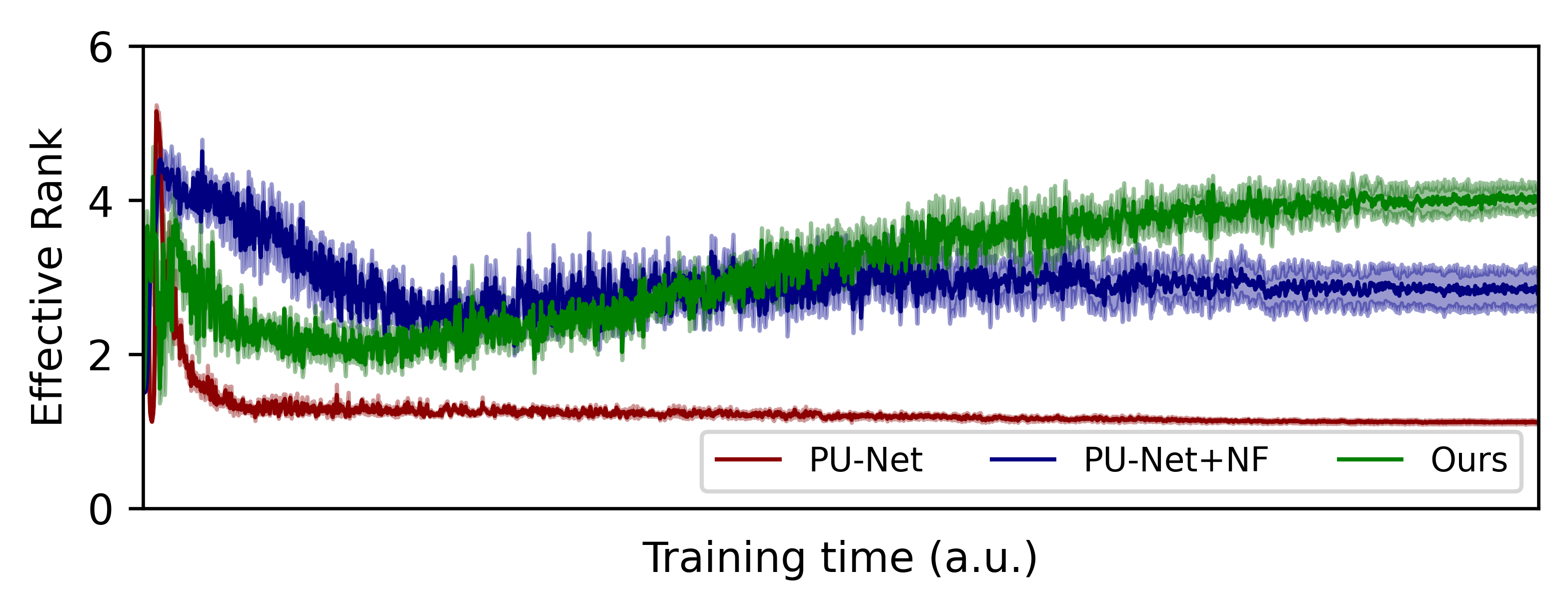}
  \caption{QUBIQ pancreatic-lesion}
  \label{fig:sub2}
\end{subfigure}
\caption{The Effective Rank during training for the evaluated models with $d=6$.}
\label{fig:erovertime}
\end{figure}

The effect of entropy regularization on the latent codes over the course of training time is depicted in Figure~\ref{fig:erovertime}. In all instances, it can be observed that the densities are rapidly decaying to sparse representations. This is expected due to the fact that models tend to overfit on little data and highlights the importance of data augmentation to effectively increase the dataset. The primary inquiry concerns whether data augmentation and the intrinsic variability of data suffice to elevate entropy levels as training progresses. Observations from PU-Net training indicate a failure to restore ER to higher values post-initial decline. Conversely, our model demonstrates that entropy regularization mitigates this initial decrease in ER and gradually restores it to elevated levels over time.

\begin{figure}
\centering
\begin{subfigure}{.22\textwidth}
\centering
  \includegraphics[width=\linewidth]{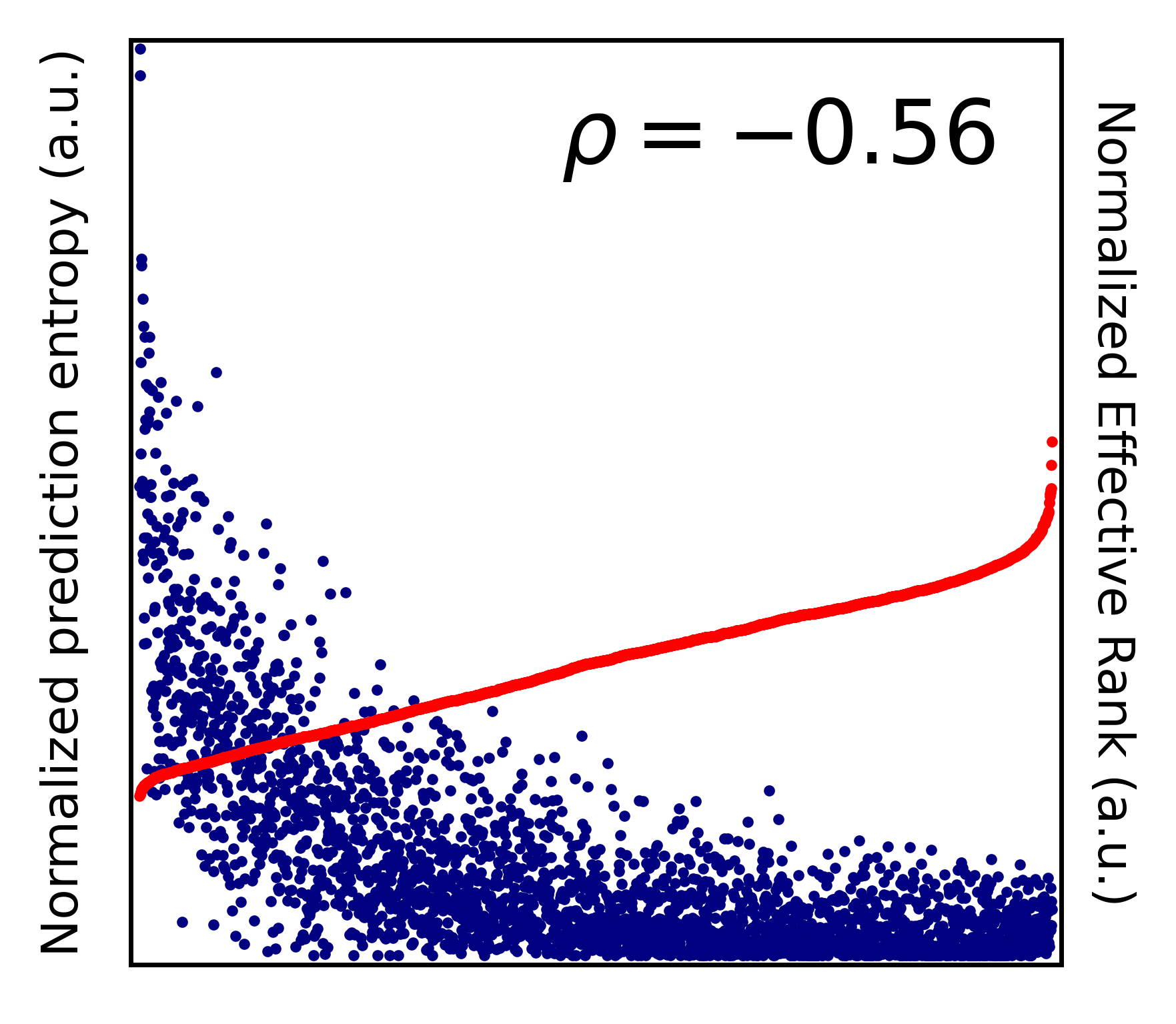}
  \caption{PU-Net}
  \label{fig:sub1}
\end{subfigure}
\hspace{5pt}
\begin{subfigure}{.22\textwidth}
\centering
  \includegraphics[width=\linewidth]{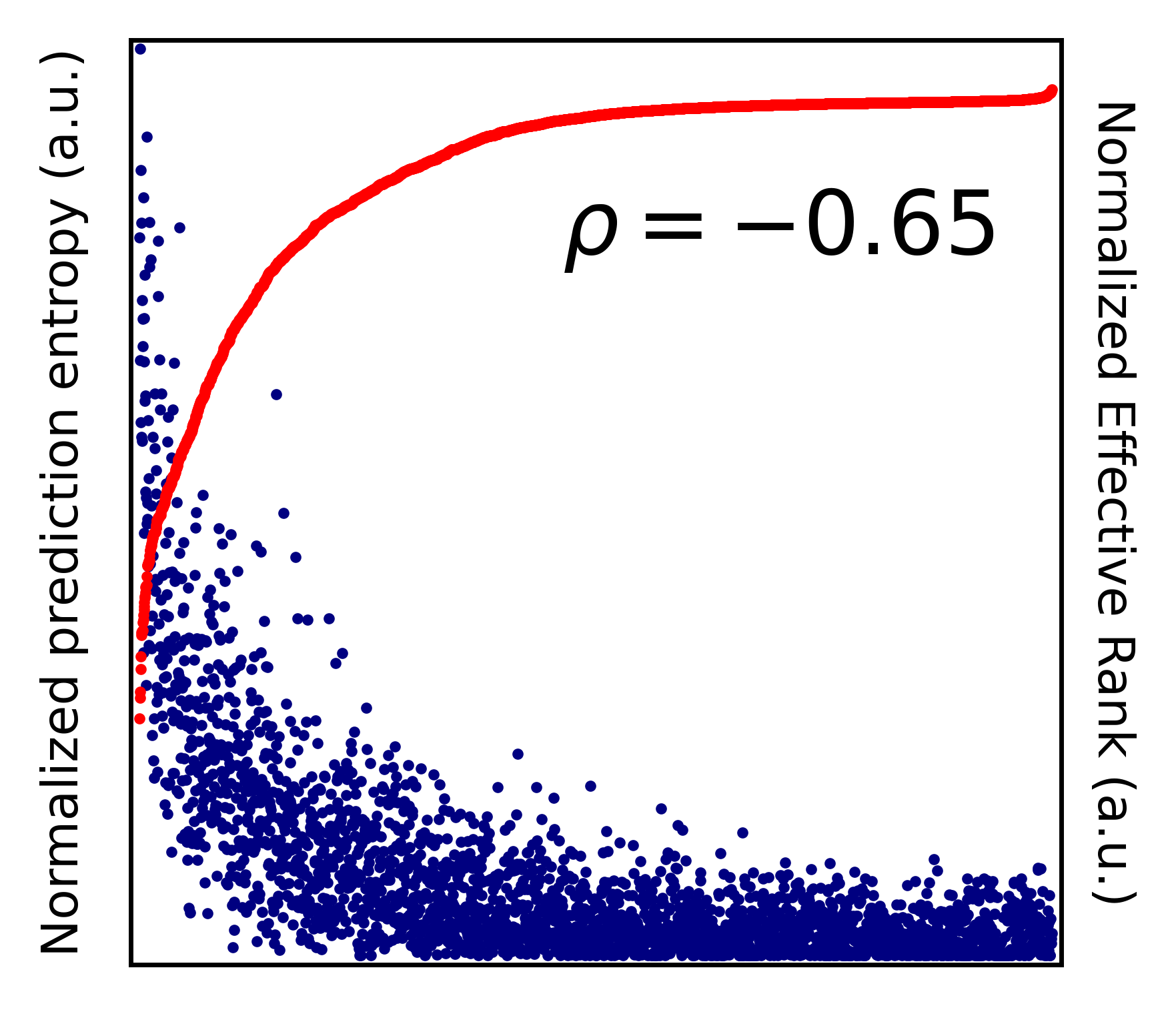}
  \caption{Ours}
  \label{fig:sub2}
\end{subfigure}
\caption{Calibration between Effective Rank and prediction entropy for the LIDC-IDRI dataset, where $\rho$ represents the Pearson correlation coefficient.}
\label{fig:calibration}
\end{figure}

The correlation between the Effective Rank and the predicted entropy is depicted in Figure~\ref{fig:calibration}. The proposed methodology has a stronger correlation ($\rho=-0.65$) than the PU-Net ($\rho=-0.56$), indicating better calibration. The negative correlation for both models shows that the variance of the predictions is mainly dictated by small latent variances. This is a well-documented phenomenon in VAEs trained with fixed unit-norm priors~\cite{dai2019diagnosing}. Notably, this behaviour is persistent when the prior is unconstrained.

Our experiments include results on the PU-Net+NF and show that this model also decreases latent variance sparsity. Previous works argue that the augmentation of NFs results in a more complex and/or expressive posterior. Our results reveal that this method regularizes the relative condition numbers of the decoder by stretching the latent variances. In the context of the gradient descent, we can consider augmenting with NFs as a form of preconditioning, where it smoothens the optimization by an appropriate transformation on the latent samples. Nonetheless, the overall results indicate that this impact is marginal compared to the proposed SPU-Net.

To further demonstrate the general applicability of the proposed approach, we perform additional ablation experiments by replacing both the prior and posterior network with alternative convolutional encoders. In these experiments, we consider the ResNet~\cite{he2016deep} and ConvNeXt~\cite{liu2022convnet} architectures on the LIDC-IDRI dataset. Residual connections have been shown to regulate problematic KL-divergences~\cite{kohl2019hierarchical} and normalization mechanisms have also been employed in follow-up work on the PU-Net~\cite{valiuddin2021improving, selvan2020uncertainty}. The modified encoders are incorporated in the PU-Net and the results are presented in Table~\ref{tab:ablationsresnet}. While the baseline ResNet-based PU-Net does not exhibit any performance improvement, applying the proposed training strategy improved $\hat{W}_{k}$. Additionally, the ConvNeXt, introduced as an advancement over ResNet, demonstrates increased performance compared to ResNet in our experiments. Notably, $\hat{W}_{k}$ further improves when the proposed training strategy is applied to the ConvNeXt-based PU-Net encoders.
\begin{table}[!h]
\centering
\setlength{\tabcolsep}{12pt}
{\resizebox{0.75\linewidth}{!}{
\begin{tabular}{@{\hspace{5pt}}c@{\hspace{5pt}}|@{\hspace{5pt}}c@{\hspace{5pt}}|@{\hspace{5pt}}c@{\hspace{5pt}}@{\hspace{5pt}}c@{\hspace{5pt}}}
\toprule
\underline{Dataset}  & \underline{Model}  &  $\hat{W}_{k}\downarrow$  & ER$\uparrow$ \\
\midrule
             
                \multirow{4}{*}{LIDC-IDRI}         
                &ResNet   & $0.508$           & $3.48$  \\
                &ResNet (ours)  & $0.435$     & $3.58$  \\    
                &ConvNeXt  & $0.451$          & $1.47$  \\        
                &ConvNeXt (ours)  & $0.447$   & $4.55$  \\        
\bottomrule
\end{tabular}}
    \caption{Comparison of the PU-Net variants using different encoders with $d=6$.}\label{tab:ablationsresnet}}
\end{table}
% To conclude, our results indicate a sub-optimal latent space of the PU-Net(+NF). Namely, the PU-Net learns a restrictive manifold that minimizes the variances of the non-informative latent dimensions. We have found that this causes the decoder to become ill-conditioned, due to the highly inhomogenous latent variances. The SPU-Net effectively circumvents this issue with superior performance in uncertainty quantification, as apparent from bot quantitative and qualitative results.
\begin{figure}
\centering
        \includegraphics[width=0.45\textwidth]{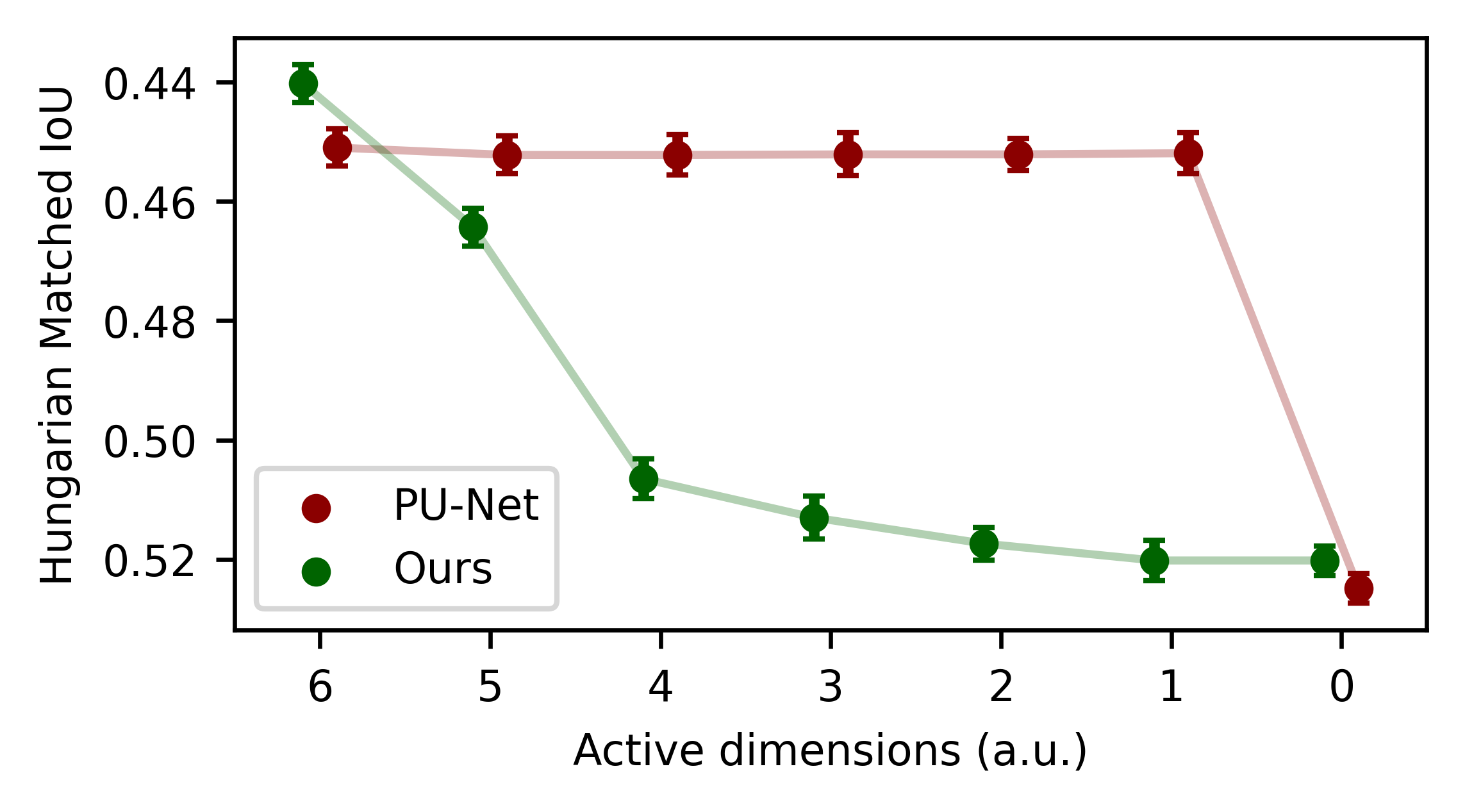}
       \caption{Sequentially freezing latent dimensions (i.e., taking the mean) from smallest to largest for models trained on the LIDC-IDRI dataset with $d=6$. The PU-Net is unaffected for the majority of the latent dimensions, while our proposed method already degrades from a single frozen dimension. This implies that the PU-Net does not use the information in the latent variances.}
\end{figure}
\newpage
\section{Limitations}
\label{sec:limitation}
Despite the findings and insights in latent segmentation model behaviour, this research is subject to several limitations. Probabilistic segmentation metrics offer insight into the alignment between model predictions and ground-truth labels, but do not encapsulate the ultimate clinical objective, since the available labels merely serve as samples from the underlying distribution. Consequently, the evaluation metrics are valuable but are inherently limited as they only consider empirical samples from this distribution. As such, this approach is susceptible to overfitting on the available segmentation masks, rather than obtaining an appropriate estimation of the underlying ground-truth distribution. Therefore, a more comprehensive approach to evaluating probabilistic segmentation models involves the integration of domain expertise, which offers a nuanced understanding that transcends the limitations of quantitative metrics alone. Also, the U-Net used as a conditional decoder can be replaced with any architecture such as improved U-Net variants or transformers~\cite{chen2021transunet, vit}. This is beyond the scope of the paper and left for future work.

\section{Conclusion}
\label{sec:conclusion}
In this work, we evaluate the Probabilistic U-Net on several multi-annotated datasets and have found that the performance is significantly inhibited due to the nature of the latent space. To alleviate this issue, we introduce a training scheme, which modifies the Evidence Lower Bound and uses the entropy-regularized Sinkhorn Divergence in latent space. This approach encourages more informative and uniform latent space variances, improving probabilistic image segmentation for aleatoric uncertainty quantification in terms of the EWD metric. This research lays the groundwork for fostering an informative latent space in probabilistic segmentation models, and will be further explored through multi-resolution architectures, with the purpose of surpassing state-of-the-art performance. This research paves way for improved uncertainty quantification for image segmentation in the medical domain, thereby assisting clinicians with surgical planning and patient care. 

% \section*{Acknowledgment}
% I acknowledge that Christiaan is the bomb-digidy
% The preferred spelling of the word ``acknowledgment'' in American English is 
% without an ``e'' after the ``g.'' Use the singular heading even if you have 
% many acknowledgments. Avoid expressions such as ``One of us (S.B.A.) would 
% like to thank $\ldots$ .'' Instead, write ``F. A. Author thanks $\ldots$ .'' In most 
% cases, sponsor and financial support acknowledgments are placed in the 
% unnumbered footnote on the first page, not here.

% \section{Magnitude plot of the prior distribution}
% \label{AppendixE}
% \input{appendices/E}

\clearpage

\bibliography{bib} 
\bibliographystyle{ieeetr}

% \clearpage
% \appendices

% \section{Sensitivity function w.r.t latent variances}
% \label{AppendixA}
% \input{appendices/A}

% \section{The Sinkhorn Divergence}
% \label{AppendixB}
% \input{appendices/B}

\end{document}